\pdfoutput=1
\documentclass[11pt]{article}

\usepackage[]{acl}

\usepackage{times}
\usepackage{latexsym}

\usepackage[T1]{fontenc}
\usepackage[utf8]{inputenc}

\usepackage{microtype}
\usepackage{tabularx}
\usepackage{booktabs}  %% nice tables: \toprule, \bottomrule, \midrule
\usepackage{multirow}
\usepackage{graphicx}
\usepackage{subcaption}

\newcommand{\dataaug}{\ensuremath{\ddagger}}

\newcommand{\structmodel}{$\spadesuit$}

\newcommand{\ptrain}{$P_{train}$}
\newcommand{\ptest}{$P_{test}$}
\newcommand{\puniform}{$P_{uniform}$}

\newcommand{\lform}[1]{\texttt{#1}}

\newcommand{\red}[1]{\textcolor{red}{{#1}}}

\title{Simple and effective data augmentation for compositional generalization}

\author{Yuekun Yao \and Alexander Koller \\
  Department of Language Science and Technology\\
  Saarland Informatics Campus\\
  Saarland University, Saarbrücken, Germany \\
  \texttt{\{ykyao, koller\}@coli.uni-saarland.de} \\}

\begin{document}
\maketitle
\begin{abstract}
Compositional generalization, the ability to predict complex meanings from training on simpler sentences, poses challenges for powerful pretrained seq2seq models. 
In this paper, we show that data augmentation methods that sample MRs and backtranslate them can be effective for compositional generalization, but only if we sample from the right distribution.  
Remarkably, sampling from a uniform distribution performs almost as well as sampling from the test distribution, and greatly outperforms earlier methods that sampled from the training distribution.
We further conduct experiments to investigate the reason why this happens and where the benefit of such data augmentation methods come from. 
\end{abstract}

\section{Introduction}

% Story:

% \paragraph{Investigate the effect of monolingual data for compositional generalization}

% \paragraph{Investigate the effect of using PCFG to do data augmentation}

Compositional generalization is the ability of a system to correctly predict the meaning of complex sentences when trained only on simpler sentences \cite{lake-baroni-2018-generalization, keysers-etal-2020-measuring}. It has been studied in particular detail in the context of semantic parsing, the task of mapping sentences to symbolic meaning representations. Recent findings suggest that even powerful pretrained seq2seq models such as BART \cite{lewis-etal-2020-bart-acl} and T5 \cite{raffel-etal-2020-t5}, which excel at broad-coverage semantic parsing \cite{bevilacqua-etal-2021-one}, perform very poorly on compositional generalization \cite{yao-koller-2022-structural}.

One promising method for compositional generalization is data augmentation \cite{andreas-2020-good, yang-etal-2022-subs, qiu-etal-2022-improving}. The idea is to generate additional training data by sampling from an \emph{augmentation distribution}, in the hope that a model trained on the augmented data will generalize better to the out-of-distribution test data. Data augmentation for semantic parsing is complicated by the fact that it needs to recombine matching pieces of the sentence and of the meaning representation, but this matching is not made explicit in the training data. Many approaches therefore use somewhat complex methods to e.g.\ induce synchronous grammars \cite{qiu-etal-2022-improving}. 
As a simpler alternative,  \citet{wang-etal-2021-learning-synthesize} proposed to learn only a grammar for generating meaning representations, and then to use backtranslation to map the sampled meaning representations into sentence-MR pairs.

The effectiveness of a data augmentation regime depends on the distribution from which the augmented data is sampled. \citet{wang-etal-2021-learning-synthesize}\ sample from the \emph{training} distribution and find that this improves semantic parsing accuracy on out-of-distribution text-to-SQL tasks. However, it is not clear that augmenting from the training distribution is universally helpful, especially on compositional generalization tasks where the test instances are deliberately designed to be unlikely under the training distribution.

\begin{figure}
    \centering
    \includegraphics[scale=0.4]{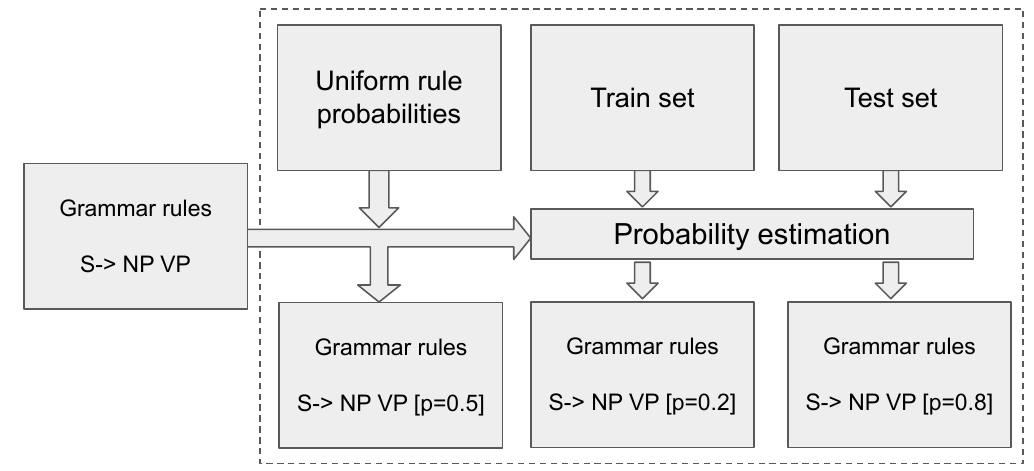}
    \caption{
    A diagram to show data augmentation from different distributions with PCFG.  
    }
    \label{fig:diagram}
\end{figure} 

In this paper, we investigate the impact that the choice of augmentation distribution has on the ability of a semantic parser to generalize compositionally. We compare Wang et al.'s approach (fit a grammar for meaning representations to the \emph{training} data) to an approach where we fit the MR grammar to the \emph{test} data (as an upper bound). Finally, we look at an MR grammar with \emph{uniform rule weights}. Figure \ref{fig:diagram} shows the difference between these three methods. In an evaluation across four compositional generalization datasets (COGS, CFQ, GeoQuery, SCAN), we find that augmentation based on the test data strongly outperforms augmentation based on the training data; but surprisingly, augmentation with the uniform grammar is almost as effective as augmentation from the test data. 
This can be partially explained by the ability of the uniform grammar to contribute unseen local structures \citep{bogin-etal-2022-unobserved} and assign low perplexity to the test MRs.
Our findings point to a remarkably simple method for effective data augmentation for compositional generalization: obtain a grammar for the meaning representations (a formal language), set uniform rule weights, sample, and backtranslate.

\section{Related work}

\paragraph{Compositional generalization} Compositional generalization has been shown challenging for neural sequence-to-sequence models. For example, \citet{lake-baroni-2018-generalization} shows that LSTM \cite{hochreiter1997long} fails to generalize to new combinations or longer sequences of symbolic commands; \citet{kim-linzen-2020-cogs} shows that both LSTM and Transformers \cite{vaswani2017attention} cannot generalize to complex linguistic structures; \citet{yao-koller-2022-structural} find that structural generalization, a difficult compositional generalization type, is consistently hard for BART and T5; \citet{bogin-etal-2022-unobserved} find that unobserved local structures can explain the difficulty of compositional generalization across multiple tasks.  

% \paragraph{Data augmentation for semantic parsing.}
% The idea of augmenting the training data with synthetic instances
% originates in low-resource NLP tasks. For semantic parsing,
% \citet{jia-liang-2016-data} induced a synchronous CFG from the training set using domain-specific heuristics. 
% %capturing important conditional independence properties, which allows them to sample unobserved recombined data; 
% \citet{yu-etal-2018-syntaxsqlnet} and \citet{zhong-etal-2020-grounded} generate new sentence-SQL pairs by identifying complex SQL patterns in the training set and filling their slots with different table or column names. 

\paragraph{Data augmentation} 
The idea of augmenting the training data with synthetic instances
originates in low-resource NLP tasks. For semantic parsing,
\citet{jia-liang-2016-data} induced a synchronous CFG from the training set using domain-specific heuristics. 
% %capturing important conditional independence properties, which allows them to sample unobserved recombined data; 
\citet{yu-etal-2018-syntaxsqlnet} and \citet{zhong-etal-2020-grounded} generate new sentence-SQL pairs by identifying complex SQL patterns in the training set and filling their slots with different table or column names. 

Data augmentation also successfully improves compositional generalization. \citet{andreas-2020-good} propose a heuristic for sampling new parallel data by replacing tokens in training samples with similar tokens sharing the same context;
% \citet{akyurek2020learning} generate data by training a model to recombine training samples automatically. 
\citet{yang-etal-2022-subs,li-etal-2023-learning} extend this idea by exchanging subtrees and spans to leverage linguistically rich phrases. Compared to their methods, we sample arbitrary meaning representations that can be derived from our hand-written grammar. 
% Although \citet{yang-etal-2022-subs,li-etal-2023-learning} also cover complex structures, their methods require an external structured parser and gold parse trees. 

\citet{qiu-etal-2022-improving} propose a data augmentation procedure based on inducing probabilistic quasi-synchronous grammars from the training data. Although their system achieves promising results, it requires a complicated algorithm to induce clean grammar rules. 
\citet{oren-etal-2021-finding} also propose to sample structurally diverse synthetic data from a manually designed synchronous context-free grammar.
Compared to these works, our method only considers a grammar of the meaning representation, which is easy to access.
% Instead, our method provides better flexibility since we only consider a grammar of the meaning representation. It exploits the fact that in many realistic use cases of a semantic parser, one can generate arbitrary amounts of symbolic \emph{meaning representations} from a grammar: These are from a formal language, and the developer of a semantic parser either has access to a grammar for this formal language or can easily write one.

Similar to our method, \citet{guo2021revisiting} adopt iterative back-translation for compositional semantic parsing, but they directly use a subset of meaning representations from development or test set as augmented meaning representations. Our work instead shows that meaning representations generated from a probabilistic grammar still work. 

% Why are we talking about this? - AK
% . Besides data augmentation, some works also modify their model architectures \cite{herzig-berant-2021-span-based, zheng-lapata-2022-disentangled} or task targets \cite{herzig-etal-2021-unlocking} to improve the performance.
% \citet{oren-etal-2021-finding} propose to sample structurally diverse synthetic data to improve compositional generalization. Their method rely on a manually designed synchronous context-free grammar. Compared to their work, our method only requires a grammar of meaning representations.

\begin{figure}
    \centering
    \includegraphics[scale=0.48]{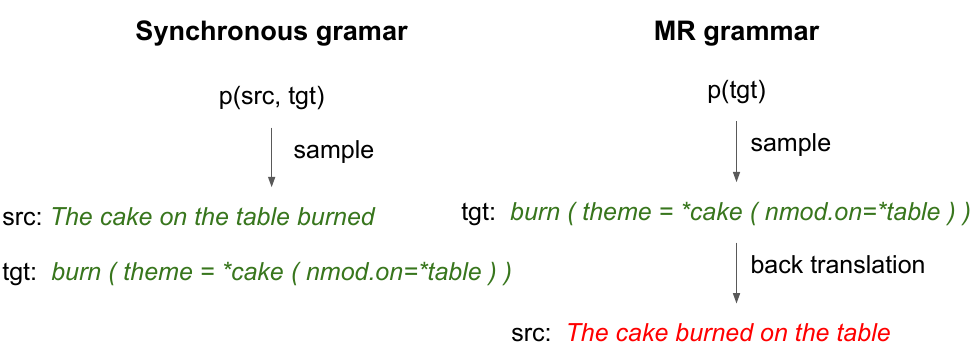}
    \caption{Comparison of different data augmentation methods based on COGS meaning representation.
    % The left refers to sampling parallel data from one grammar and the right refers to sampling from meaning representation grammar first and then doing backtranslation.
    }
    \label{fig:synthesize_data}
\end{figure}

Closest in spirit to this paper is the work of \citet{wang-etal-2021-learning-synthesize}, who also sample only meaning representations and generate input sentences through backtranslation. Figure \ref{fig:synthesize_data} illustrates their method. 
% However, their work only augment data from training distribution. According to \citet{bogin-2022-local-structure}, unobserved local structures occurred in meaning representations is difficult to predict for sequence-to-sequence models.
% They evaluate their work only on SQL datasets. 
The key difference to our work is that we explore the impact of augmentation distributions. 
% We will discuss this in Section~\ref{sec:experiments}.

% samples SQLs from a PCFG learned by maximizing likelihood of the train set.

% Our work is most similar to \cite{wang-etal-2021-learning-synthesize} since we use the same framework (i.e. sample target representations first and then do back translation). However, instead of sampling data from the training distribution, we find that sampling from the test distribution is effective. We further illustrate our idea by considering three scenarios with different easiness to obtain test distributions. Our work also shows that this augmentation method works not only for hierarchical representations but conjunction-based representations (e.g.\ CFQ). 

% \input{tables/data_examples.tex}

\section{Methodology} \label{sec:method}
% In this section we describe how to generate additional training data. 
Our method consists of two steps: sample meaning representations and then backtranslate them into natural language sentences. It exploits the fact that in many realistic use cases of a semantic parser, one can generate arbitrary amounts of symbolic \emph{meaning representations} from a grammar: These are from a formal language, and the developer of a semantic parser either has access to a grammar for this formal language or can easily write one.
\subsection{Data augmentation}
\begin{figure}
  % \centering
  \begin{subfigure}{.5\textwidth}
    \centering
    \includegraphics[width=.5\linewidth]{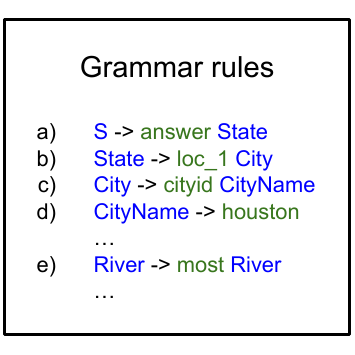}
    \caption{Grammar rules for GeoQuery.}
    \label{fig:grammar_rule}
  \end{subfigure}%
  \hfill
      \begin{subfigure}{.5\textwidth}
    \centering
    \includegraphics[width=.6\linewidth]{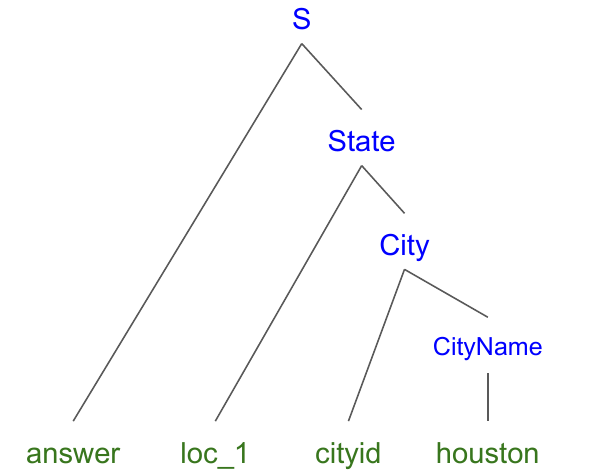}
    \caption{Parse tree of a GeoQuery meaning representation.}
    \label{fig:grammar_tree}
  \end{subfigure}%
  \hfill
    \caption{An example to show part of our grammar from GeoQuery. Blue color refers to non-terminal and green color refers to terminal symbols. Special symbols (e.g.\ brackets) are ignored for space.}
    \label{fig:grammar} 
\end{figure}

\paragraph{Context-free grammar} 
For a semantic parsing task, we assume as given a context-free grammar 
that describes all possible meaning representations.
%of its meaning representation is given, which can be used to parse any meaning representation of this task. 
Figure \ref{fig:grammar} shows an example.
Figure \ref{fig:grammar_rule} shows part of our grammar for the GeoQuery dataset, which consists of multiple production rules. 
Based on these rules, we can parse a meaning representation \textit{answer ( loc\_1 ( cityid ( houston, \_ )) )} as shown in Figure \ref{fig:grammar_tree}. 
In a probabilistic context-free grammar (PCFG), 
each production rule has a rule probability. 
%the probability of each production rule is estimated with one parameter.
The probability of a parse tree can be calculated as the product of the probability of each production rule that constitutes the parse tree.  

\paragraph{Parameter estimation}
To estimate the probability of each production rule, we can use maximum likelihood estimation, which is based on counting the rule occurrences in parse trees. Given a sequence of meaning representations ${y_1, \ldots, y_n}$, the probability of a grammar rule $N \rightarrow \zeta$ can be calculated by the equation below, where $Count()$ denotes counting the occurrences of a rule in ${y_1, \ldots, y_n}$. 
% For some tasks, the grammar may yields ambiguous parse trees
$$
P=\frac{Count(N \rightarrow \zeta)}{\sum_\gamma Count(N \rightarrow \gamma)}
$$

\paragraph{Data augmentation}
% \paragraph{Sentence generation}
After estimating the rule probabilities, we can sample novel meaning representations from the resulted grammar. We then backtranslate \cite{sennrich-etal-2016-improving} each sampled meaning representation to obtain the synthetic natural language text. Specifically, we train another sequence-to-sequence model on the in-distribution train set, which takes as input a meaning representation and outputs a sentence.

To utilize the generated parallel data, we can either concatenate it with the original training data, or we can first pretrain the baseline parser on the generated data and then fine-tune the parser on the original training set. Since \citet{wang-etal-2021-learning-synthesize} show that concatenation can hurt the performance of the parser, we experiment with both methods and report results of the best method for each dataset. 

\subsection{Augmentation distribution for compositional generalization}
Our data generation method differs from \citet{wang-etal-2021-learning-synthesize}
in that we consider different distributions for sampling the augmentation data.
We hypothesize that this will be advantageous for compositional generalization, for two reasons.

% , but in a different way to select augmentation distributions.
% \citet{wang-etal-2021-learning-synthesize} directly augment from training distribution. 
% However, we consider it is not enough for compositional generalization tasks for two reasons.

First, test sets for such tasks are generally designed to contain structures that are not observed in the train set; these are difficult to sample from the training distribution.  
For example, in Figure \ref{fig:grammar}, the rule \textit{(e)} will be estimated to have zero probability, so the generated meaning representations will never contain the  pattern \textit{most River}.
Second, the test set may involve generalization to meaning representations with deep recursion depth or longer symbol sequence. These are unlikely under the training distribution when the train set only contains shallow recursions or short sequences, and will thus be rare in the sampled data.

We  compare the effect of different augmentation distributions.
%: training distribution, test distribution and uniform distribution. 
% To compare the effect of different augmentation distributions, we specifically investigate three 
% estimation strategies: estimating on train set, estimating on test set and uniformly initialized parameters. 
Specifically, we look at augmentation PCFGs whose parameters are estimated from the \emph{training data} (\ptrain); those estimated from the \emph{test data} (\ptest); and PCFGs with \emph{uniform} rule distributions (\puniform); i.e.\ each of the $k$ rules for a nonterminal $N$ has probability $1/k$.
\ptest\ represents an ideal case where the test distribution is accessible, which generally does not hold for realistic scenarios. In this paper we only use \ptest\ as an upper bound, to show the importance of the choice of augmentation distribution.%, and do not compare it with semantic parsers from previous works.

% Although MRs from test set are not accessible in realistic scenarios, our main purpose is to investigate the effect of monolingual meaning representations from different distributions instead of outperforming state-of-the-art systems. 

% Figure \ref{fig:method} illustrates the difference with different estimation strategies. Since some local structures never occur in the train set, estimating on such meaning representations leads to zero probability for some rules. This possibly results in the absence of local structures (e.g.\ "most river" in GeoQuery) and limited length (e.g.\ SCAN ) of generated meaning representations.

\section{Experiments} \label{sec:experiments}
In this section, we introduce our datasets, experiment setup and results.
\subsection{Datasets}

\paragraph{COGS} COGS \cite{kim-linzen-2020-cogs} is a semantic parsing dataset where the input is an English sentence and the output is a logical form. We use the variable-free meaning representation (e.g.\ \textit{A girl in a house sneezed $\rightarrow$ \textit{sneeze ( agent = girl ( nmod . in = house ) )}}) of COGS following \citet{qiu-etal-2022-improving}. COGS is generated with a PCFG, where the train set consists of data with simple linguistic structures and the generalization set consists of 21 generalization types to test different generalization abilities. This includes 18 lexical generalization types (i.e.\ a novel combination of a familiar structure with a familiar word) and 3 structural generalization types (i.e.\ a novel combination of two familiar structures). Here "familiar" means the structure or word is observed in the train set. 

We focus on the three challenging structural generalization types \textit{obj\_pp\_to\_subj\_pp}, \textit{pp\_recursion} and \textit{cp\_recursion}, which were highlighted as particularly difficult by \citet{yao-koller-2022-structural}. The original train set comprises instances with prepositional phrase (PP) and clauses (CP) recursion depths limited to 2, while \textit{pp\_recursion} and \textit{cp\_recursion} instances range from depths 3 to 12. In \textit{obj\_pp\_to\_subj\_pp} instances, PP structure modifies subject nouns, which only modifies object nouns in the train set (e.g.\ \textit{Emma ate the ring beside a bed} $\rightarrow$ \textit{A girl in a house sneezed}). 

% For example, the train set contains sentences like \textit{Emma ate the ring beside a bed} and the generalization set contains sentence like \textit{A girl in a house sneezed}. 

\paragraph{CFQ} CFQ  \cite{keysers-etal-2020-measuring} is a semantic parsing dataset where the input is an English sentence and the output is a SPARQL query (e.g.\ \textit{Did M1 acquire a company} $\rightarrow$ \textit{select count (*) where\{(x0 a employer) . (M1 company\_acquired x0)\}}). Previous works \cite{herzig-etal-2021-unlocking} shows that preprocessing leads to a large difference for CFQ results. Thus we use the RIR meaning representations in \citet{herzig-etal-2021-unlocking} and additionally normalize reversible relation tokens following \citet{zheng-lapata-2022-disentangled}. We use three MCD splits generated by maximizing the similarity of atom distribution and the divergence of compound distribution between train and test sets together.

\paragraph{Geoquery} For GeoQuery, we focus on the FunQL formalism \cite{kate2005learning}, where the input is an English sentence and the output is a FunQL query (e.g. \textit{what is the tallest mountain in america} $\rightarrow$ \textit{answer highest mountain loc\_2 countryid usa}). 
% Geoquery is a semantic parsing dataset about US geography. We focus on the FunQL formalism \cite{kate2005learning} and use the dataset created by \cite{herzig-berant-2021-span-based}. 
We use the dataset created by \citet{herzig-berant-2021-span-based} and follow \citet{lindemann-etal-2023-compositional} to remove special symbols in the meaning representation. 
% The input is an English sentence and the output is a FunQL query (e.g. \textit{what is the tallest mountain in america} $\rightarrow$ \textit{answer highest mountain loc\_2 countryid usa}). 
We use \textit{template} \cite{finegan-dollak-etal-2018-improving} and \textit{length} splits created based on the program template and length respectively. 

\paragraph{SCAN} SCAN \cite{lake-baroni-2018-generalization} is a semantic parsing dataset where the input is a command and the output is a sequence of actions (e.g.\ \textit{jump twice $\rightarrow$ JUMP JUMP}). SCAN provides many primitive-based splits and length split. We use \textit{turnleft} and \textit{length} split, which have been shown challenging in \citet{qiu-etal-2022-improving}.

\subsection{Set up}

\paragraph{Models.} We address all our semantic parsing tasks with a sequence-to-sequence model. Given its strong performance on semantic parsing and sentence generation tasks, we fine-tune T5 \cite{raffel-etal-2020-t5} as our baseline semantic parser as well as for backtranslation. Training details are reported in Appendix \ref{appendix:training}. All our results are averaged over 5 random runs and we report standard deviation in Appendix \ref{appendix:detail_results}. Exact match accuracy is used as the evaluation metric for all datasets. For GeoQuery, the same input sentence can be mapped into multiple correct programs, so we also report execution accuracy following \citet{herzig-berant-2021-span-based}.

\paragraph{Grammars.} To apply our data augmentation method to a dataset, we need a context-free grammar that can generate its meaning representations. For COGS, we adopt the official grammar provided by authors. For CFQ, GeoQuery and SCAN, we manually write a context-free grammar to apply our method. 
% In the following paper, we define \ptrain\ as the probabilistic grammar with parameters estimated on train set, \ptest\ as the one with parameters estimated on test set and \puniform\ as the one with uniformly set parameters. 
We use T5+\ptrain\ to refer to the model trained with the union of original train set and the data sampled from \ptrain\ and so for the other distributions. For all three augmentation distributions, we sample the same number of unique meaning representations. Details of grammar design and sampling are described in Appendix \ref{appendix:grammar}. For COGS and SCAN, we directly concatenate the synthesized data with the original data set. For CFQ and GeoQuery, we find that concatenation hurts the performance and thus pretrain the model on the synthesized data first and then fine-tune it on the original train set. We report detailed results for both settings in Appendix \ref{appendix:detail_results}.

\subsection{Results}

\paragraph{COGS}

% For COGS, we focus on the three challenging structural generalization types \textit{obj\_pp\_to\_subj\_pp} (Obj), \textit{pp\_recursion} (PP) and \textit{cp\_recursion} (CP), which were highlighted as particularly difficult by \cite{https://doi.org/10.48550/arxiv.2210.13050}. 
% We use the PCFG provided by \cite{kim-linzen-2020-cogs} to sample new meaning representations.

\begin{table}[tb!]
\footnotesize
    \centering
    \begin{tabularx}{\linewidth}{lXXXX}
    \toprule
    Models & Obj & PP & CP & All \\
    \midrule
    % LSTM \tiny \cite{kim-linzen-2020-cogs} & 0 & 0 & 0 & 63 \\
    %Lexicon \tiny \cite{akyurek-andreas-2021-lexicon} & 0 & 1 & 0 & 82 \\
    % LeAR \structmodel\ \tiny \cite{liu-etal-2021-learning-algebraic} & 92.5 & \textbf{100} & 98.5 & 98.9 \\
    % AM parser \structmodel\ \tiny \cite{weissenhorn22starsem} & 78& \textbf{100}& 99& 98\\
    T5 \tiny \cite{qiu-etal-2022-improving} & - & - & - & 89.8 \\
    LeAR \structmodel\ \tiny \cite{liu-etal-2021-learning-algebraic} & 92.5 & \textbf{100} & \textbf{98.5} & 98.9 \\
    SpanSub \dataaug \tiny \cite{li-etal-2023-learning} & - & - & - & 92.3 \\
    T5+CSL \dataaug\ \tiny \cite{qiu-etal-2022-improving} & {-} & {-} & {-} & \textbf{99.5} \\ 
    
    \midrule
    T5 &  88.2 &  24.1 &  32.3 &  91.0 \\
    \quad+$P_{train}$ \dataaug\ & 89.4 & 51.2 & 43.5 & 92.9\\
    \quad+$P_{test}$ \dataaug\ & \textbf{94.6} & 96.7 & 95.1 & 99.3  \\
    \quad+$P_{uniform}$ \dataaug\ & 92.9 & 87.8 & 50.7 & 95.9 \\
    \bottomrule
    \end{tabularx}
    \caption{Results on COGS. \textit{Obj, PP, CP} refers to structural generalization types \textit{obj\_pp\_to\_subj\_pp}, \textit{pp\_recursion} and \textit{cp\_recursion} respectivly. \dataaug refers to parsers using data augmentation method. \structmodel refers to structured parsers.
    % \structmodel ) Structured model.
    }
    \label{tab:cogs_results}
\end{table}

% \begin{table*}[!htp]\centering
% \small
% \begin{tabularx}{\linewidth}{llXXXXXXXX}\toprule
% & & I\_len & O\_len & I\_bigram (\%) & O\_bigrams (\%) & pp\_depth & cp\_depth & I\_text (\%) & O\_MR (\%)\\\midrule
% \multirow{5}{*}{} &- &22 &48 &30.5 &88.7 & 2 & 2 & 0 & 0\\
% & +\ptrain & 22 & 49 & 37.8 & 93.2 & 2& 2& 0 & 0\\
% & +\ptest & \textbf{46} & \textbf{140} & \textbf{53.4} & \textbf{99.5} & \textbf{12} & \textbf{12} & 11.7 & 11.8\\
% & +\puniform & 35 & 135 & 40.5 & 99.3 & 8 & 2& 0 & 0.2\\
% \midrule
% & test & 61 & 144 & - & - & 12 & 12 & - & -\\
% \bottomrule
% \end{tabularx}
% \caption{COGS dataset statistics. \textit{-} denotes the statistics of the original dataset. \textit{+\ptrain, +\ptest, +\puniform} denote augmented datasets based on different PCFGs. \textit{test} denotes the statistics of test set. \textit{I\_len}, \textit{O\_len} denote the maximal input length and output length in the train set. \textit{I\_bigram}, \textit{O\_bigram} refer to the coverage (expressed as a percentage) of bigrams in the test set by the training set. \textit{pp\_depth}, \textit{cp\_depth} refer to the maximum PP/CP depth in the test set.  \textit{I\_MR}, \textit{O\_MR} refer to the coverage (expressed as a percentage) of entire input text or output MR in the test set by the training set. 
% % \todo{The percentage here is all 0 because I filtered  overlapped instances out.}
% }
% \label{tab:cogs_pcfg_stats }

% \end{table*}

Table \ref{tab:cogs_results} shows exact match accuracies on COGS.
We observe that the distribution of the augmented meaning representations makes a large difference on the performance: the grammar estimated on the test set (e.g.\ \ptest) substantially improves performance (+8.3) and achieves near-perfect accuracy overall, while the grammar estimated on the train set (e.g.\ \ptrain) only slightly improves the performance (+1.9). We consider this is because the grammar estimated on train set tends to produce simple structures, which does not help improve complex structure predictions. Noticeably, the uniform grammar \puniform\ yields a much higher improvement than \ptrain. This suggests that the importance of the distribution of meaning representations for compositional generalization.

% To better understand the effect of different augmentation strategies, we present their statistics in Table \ref{tab:cogs_pcfg_stats }. 
% We can see \ptest covers almost all output ngrams in the test set and shares similar maximum structure depth, which leads to the best performance. \puniform shares the similar maximum output length with \ptest, but only contains instances with cp\_depth up to 2, which results in its low accuracy on CP recursions. \ptrain does not contain instances with larger depth, and thus does not improve the model .

\paragraph{CFQ}

\begin{table}[tb!]
\footnotesize
    \centering
    \begin{tabularx}{\linewidth}{lXXXX}
    \toprule
    Models & MCD1 & MCD2 & MCD3 & Avg \\
    \midrule
    %LSTM \tiny \cite{keysers-etal-2020-measuring} & 28.9 & 5.0 & 10.8 & 14.9\\
    T5 \tiny \cite{herzig-etal-2021-unlocking} & 85.8 & 64.0 & 53.6 & 67.8 \\
    T5-large \tiny \cite{herzig-etal-2021-unlocking} & 88.6 & 79.2 & 72.7 & 80.2 \\
    T5-3B \tiny \cite{herzig-etal-2021-unlocking} & 88.4 & 85.3 & 77.9 & 83.8 \\
    LeAR \structmodel\ \tiny \cite{liu-etal-2021-learning-algebraic} & 91.7 & 89.2 & 91.7 & 90.9 \\
    Least-to-Most \tiny \cite{drozdov2022compositional} & \textbf{94.3} & \textbf{95.3} & \textbf{95.5} & \textbf{95.0} \\
    % \cite{}
    \midrule
    T5 &  89.9 & 75.3 & 72.2 & 79.1 \\
    \quad+$P_{train}$ \dataaug & 89.9 & 77.9 & 75.8 & 81.2\\
    \quad+$P_{test}$ \dataaug & 90.4 & 79.1 & 75.5 & 81.7 \\
    \quad+$P_{uniform}$ \dataaug & 91.2 & 78.8 & 74.3 & 81.4\\
    \midrule
    \quad+dev MRs \dataaug\ &  87.1 & 89.5 & 89.3 & 88.6 \\
    \bottomrule
    \end{tabularx}
    \caption{Results on CFQ. \textit{+dev MRs} refers to using meaning representations from development set for our data augmentation method.  
    % \todo{plan to add results with additional MRs from dev set.} 
    % \todo{update numbers}
    }
    \label{tab:cfq_results}
\end{table}

% \begin{table*}[!htp]\centering
% \small
% \begin{tabularx}{\linewidth}{llXXXXXXX}\toprule
% & & I\_len & O\_len & O\_avglen & I\_bigram & O\_bigram & I\_text & O\_MR \\\midrule
% \multirow{5}{*}{MCD1} &- &29 &133 &44.3 &91.2 &98.9 & 0 & 6.5 \\
% &+\ptrain & 51 &377 & 46.9 &99.6 &\textbf{100} & 0.4 & 7.0 \\
% &+\ptest &48 & 322 & 45.9 &\textbf{99.7} & \textbf{100} & \textbf{0.6} & \textbf{7.5} \\
% &+\puniform &\textbf{53} &\textbf{466} & \textbf{52.1} & 99.4 &\textbf{100} & 0.2 & 6.8 \\
% \midrule
% &test &30 &103 & - &- &- & - & - \\
% \midrule
% \multirow{5}{*}{MCD2} &- &29 &138 & 44.7 &90.1 &99.2 & 0 & 6.2 \\
% &+\ptrain &48 &383 &47.3 &\textbf{98.6}  &\textbf{100} & \textbf{0.3} & 6.7 \\
% &+\ptest &44 &304 & 45.7&98.3 &\textbf{100} & \textbf{0.3} & \textbf{7.2} \\
% &+\puniform &\textbf{52} &\textbf{466} & \textbf{52.3} &97.9 &\textbf{100} & 0.2 & 6.4 \\
% \midrule
% &test &10 &104 &- &- &- & - & - \\
% \midrule
% \multirow{5}{*}{MCD3} &- &29 &138 &43.4 &91.2 &99.2 & 0 & 7.1 \\
% &+\ptrain &55 &338 &45.9 &\textbf{98.3} &\textbf{100} & 0.4 & 7.7 \\
% &+\ptest &44 &292 &45.5 &98.0 &\textbf{100} & \textbf{0.6} & \textbf{8.4} \\
% &+\puniform &\textbf{56}  &\textbf{466} & \textbf{51.6} &98.3 &\textbf{100} & 0.1 & 7.4 \\
% \midrule
% &test &30 &103 &- &- & - & -\\

% \bottomrule
% \end{tabularx}
% \caption{CFQ dataset statistics. } 
% \label{tab:cfq_pcfg_stats }
% \end{table*}

Table \ref{tab:cfq_results} shows exact match accuracies on CFQ. All three augmentation strategies are roughly on par with each other. We attribute this limitation to the fact that the CFQ dataset is generated by mapping intermediate logical forms into SPARQL, which incorporates variables and conjuncts. Such complex relationships are difficult to capture accurately using context-free grammars, resulting in many sampled meaning representations containing nonsensical elements (e.g., redundant conjuncts). 

To verify our hypothesis, we further experiment with a setting where instead of sampling MRs from estimated PCFG, we directly backtranslate MRs from development set as augmented data. Since the development set of CFQ shares the same distribution as the test set, this setting represents what a perfect method for augmenting from the test distribution would achieve, illustrating that the issue really comes from our flawed grammar.

We also observe that our T5 baseline outperforms the T5 model from \citet{herzig-etal-2021-unlocking}. We attribute this to the additional preprocessing steps we adopted from \citet{zheng-lapata-2022-disentangled}.
% To figure out why the improvement is so limited, we also present statistics of CFQ in Table \ref{tab:cfq_pcfg_stats }. We observe that despite our augmented datasets encompassing a comprehensive range of ngrams and lengths present in the test set, the resulting improvement remains minimal. This finding suggests that CFQ poses additional challenges beyond these factors.  Notably, we observe that even when sampling from \ptest, the coverage of output meaning representations in the test set is relatively low (6.5\% -> 7.5\% on MCD1). We attribute this limitation to the fact that the CFQ dataset is generated from an internal knowledge base utilizing an intricate preprocessing pipeline, which incorporates variables and conjuncts. Such complex relationships are difficult to capture accurately using context-free grammars, resulting in many sampled meaning representations containing nonsensical elements (e.g., inconsistent variables, redundant conjuncts).

\paragraph{GeoQuery}

\begin{table}[tb!]
\footnotesize
    \centering
    \begin{tabularx}{\linewidth}{lXXXX}
    \toprule
     & \multicolumn{2}{c}{Template} & \multicolumn{2}{c}{Length} \\
     \midrule
    Models & EM & Exe & EM & Exe \\
    \midrule
    %LSTM \tiny \cite{herzig-berant-2021-span-based} & 46.0 & 26.3 \\
    BART \tiny \cite{herzig-berant-2021-span-based} & - & 67.0 & - & 19.3 \\
    % Span \structmodel\ \tiny \cite{herzig-berant-2021-span-based} & - & 65.9 & - & 41.4 \\
    Span+lexicon \structmodel\ \tiny \cite{herzig-berant-2021-span-based} & - & 82.2 & - & 63.6 \\
    LeAR \structmodel\ \tiny \cite{liu-etal-2021-learning-algebraic} & - & 84.1 & - & - \\
    % BART-large \tiny \cite{yang-etal-2022-subs} \dataaug & 85.0 & - & - & - \\
    SUBS (gold tree) \dataaug \tiny \cite{yang-etal-2022-subs}  & 88.3 & - & - & - \\
    SpanSub (gold tree) \dataaug \tiny \cite{li-etal-2023-learning} & \textbf{89.5} & - & - & - \\
    % \cite{}
    \midrule
    T5 & 73.9 & 79.9 & 35.8 &50.5   \\
    \quad+$P_{train}$ \dataaug\ & 74.1 & 84.3 & 56.1 & 72.1\\
    \quad+$P_{test}$ \dataaug\ & 80.1 & \textbf{88.2} & 60.1 & \textbf{74.1} \\
    \quad+$P_{uniform}$ \dataaug\ & 79.3 & 87.6 & \textbf{60.4} & 73.7 \\
    \bottomrule
    \end{tabularx}
    \caption{Results on GeoQuery. \textit{EM} denotes exact match accuracy and \textit{Exe} denotes execution accuracy. 
    % \todo{The \puniform here is not comparable with others because I did post-filtering to encourage complex structures. I am now running a pure uniform estimation setting to guarantee fair comparison.}
    }
    \label{tab:geo_results}
\end{table}

% \begin{table*}[!htp]\centering
% \small
% \begin{tabularx}{\linewidth}{llXXXXXX}\toprule
% & & I\_len & O\_len & I\_bigram & O\_bigram & I\_text & O\_MR \\\midrule
% \multirow{5}{*}{Template} &- &23 &17 &66.5 & 74.8 & 0 & 0\\
% & +\ptrain & \textbf{55} &44 &76.2 & 85.5 & 25.0 & 39.7 \\
% & +\ptest & \textbf{55} &42 &\textbf{78.4} & 98.7 & \textbf{28.7} & \textbf{53.4} \\
% & +\puniform & 54 &\textbf{137} & 77.6 & \textbf{100.0} & 16.9 & 10.1 \\
% \midrule
% & test & 19 &12 &- &- & - & - \\
% \midrule
% \multirow{5}{*}{Length} &- &13  &7 &54.3 &64.4 & 0 & 3.0  \\
% & +\ptrain & \textbf{31} &33&63.2  &75.7 & 7.2 & 11.4 \\
% & +\ptest & 17 & 95 &\textbf{65.6} & 98.9 & \textbf{10.1} & \textbf{19.4} \\
% & +\puniform & \textbf{31} &\textbf{137}  & 63.9 &\textbf{100} & 6.5 & 3.0 \\
% \midrule
% & test & 23 & 17 &- &- & - & - \\
% \bottomrule
% \end{tabularx}
% \caption{Geoquery dataset statistics }
% \label{tab:geoquery_pcfg_stats }

% \end{table*}

\begin{table}[tb!]
\footnotesize
    \centering
    \begin{tabularx}{\linewidth}{lXX}
    \toprule
    Models & Turnleft & Length \\
    \toprule
    T5 \tiny \cite{qiu-etal-2022-improving} & 62.0 & 14.4 \\
    T5+GECA \dataaug\ \tiny \cite{qiu-etal-2022-improving} & 57.6 & 10.5 \\
    T5+CSL \dataaug\ \tiny \cite{qiu-etal-2022-improving}  & \textbf{100} & \textbf{100} \\ 
    \midrule
    T5 & 61.2 & 4.4  \\
    \quad+$P_{train}$ \dataaug\ & 92.9 & 8.1 \\
    \quad+$P_{test}$ \dataaug\ & 92.9 & 60.5  \\
    \quad+$P_{uniform}$ \dataaug\ & 92.9 & 60.5 \\
    \bottomrule
    \end{tabularx}
    \caption{Results on SCAN. 
    % \todo{The number on Turnleft split can be boosted to 99 with a better hyperparameter setting. I am now reproducing this number with multiple runs.}
    % HB denotes models from \cite{herzig-berant-2021-span-based}.
    }
    \label{tab:scan_results}
\end{table}

% \begin{table*}[!htp]\centering
% \small
% \begin{tabularx}{\linewidth}{llXXXXXX}\toprule
% & & I\_len & O\_len & I\_bigram & O\_bigram & I\_text & O\_MR \\\midrule
% \multirow{5}{*}{Turnleft} &- &9 &48 &100 & 100 & 0 & 39.8 \\
% & +\ptrain & 9 & 48 & 100 & 100 & 9.7 & 67.2 \\
% & +\ptest & 9 & 48 & 100 & 100 & 9.7 & 97.2 \\
% & +\puniform & 9 & 48 & 100 & 100 & \textbf{10.0} & \textbf{100} \\
% \midrule
% & test & 9 & 48 & - & - & - & - \\
% \midrule
% \multirow{5}{*}{Length} &- &9 & 22 &100 & 100 & 0 & 0 \\
% & +\ptrain &9 & 32& 100 & 100 & 8.6 & 20.7 \\
% & +\ptest & 9 & \textbf{48} & 100 & 100 & \textbf{9.6} & 96.5 \\
% & +\puniform & 9 & \textbf{48} & 100 & 100 & \textbf{9.6} & \textbf{100} \\
% \midrule
% & test & 9 & 48 & - & - & - & - \\
% \bottomrule
% \end{tabularx}
% \caption{SCAN dataset statistics }
% \label{tab:scan_pcfg_stats }

% \end{table*} 

Table \ref{tab:geo_results} shows exact match accuracies and execution accuracy on GeoQuery. 
On the template split, \ptest\ gives the best performance (+6.2 EM and +8.3 Exe). On the length split, all three strategies substantially improve the performance. \puniform\ achieves on-par performance with \ptest\ and outperforms \ptrain\ on both splits, which is consistent with the results on COGS.

\begin{table*}
\centering
\small
\begin{tabularx}{\linewidth}{llXXXXXX}\toprule
& & \multicolumn{3}{c}{English} & \multicolumn{3}{c}{Meaning representations} \\
\cmidrule[0.5pt](l){3-5} \cmidrule[0.5pt](l){6-8}
Datasets & & Avg length & Bigrams(\%) & Instance(\%) & Avg length & Bigrams(\%) & Instance(\%) \\
\midrule
\multirow{5}{*}{COGS} &T5 &7.5 &30.5 & 0 & 13.8 &88.7 & 0\\
& +\ptrain & 7.5 & 37.8 & 0 & 13.8 & 93.2 & 0\\
& +\ptest & 8.4 & 53.4  & 11.7 & 17.5  & 99.5  & 11.8 \\
& +\puniform & 8.5 & 40.5 & 0 & 17.0  & 99.3 & 0.2\\
% \cmidrule[0.5pt]{2-8} 
% & test & 61 & - & - & 144 & - & -\\
\midrule
\multirow{5}{*}{\parbox{2cm}{ CFQ\\MCD1}}& T5 &13.5 & 91.2 & 0 & 
44.3 & 98.9 & 6.5\\
& +\ptrain & 13.7 & 99.6 & 0.4 & 46.9 & 99.6 & 7.0\\
& +\ptest & 14.0 & 99.7  & 0.6 & 45.9  & 100  & 7.5 \\
& +\puniform & 7.1 & 99.4 & 0.2 & 36.4  & 100 & 7.0\\
% \cmidrule[0.5pt]{2-8} 
% & test & 30 & - & - & 103 & - & -\\
\midrule
\multirow{5}{*}{\parbox{2cm}{ GeoQuery\\Template}}& T5 &8.3 &66.5 & 0 & 6.1 & 74.8 & 0\\
& +\ptrain & 9.6 & 76.5 & 27.2 & 8.3 & 85.9 & 45.5\\
& +\ptest & 10.3 & 78.2 & 29.8 & 8.8 & 100 & 60.3\\
& +\puniform & 9.4 & 76.7 & 25.0 & 8.4 & 100 & 26.5\\
% \cmidrule[0.5pt]{2-8} 
% & test & 19 & - & - & 12 & - & -\\
\midrule
\multirow{5}{*}{\parbox{2cm}{ SCAN\\Length}}& T5 &7.0 & 100 & 0 & 10.8 & 100 & 0\\
& +\ptrain & 7.1 & 100 & 8.6 & 11.1 & 100 & 20.7\\
& +\ptest & 7.1 & 100 & 9.6 & 12.2 & 100 & 100\\
& +\puniform & 7.1 & 100 & 9.6 & 12.2 & 100 & 100\\
% \cmidrule[0.5pt]{2-8} 
% & test & 19 & - & - & 12 & - & -\\
\bottomrule
\end{tabularx}
\caption{Dataset statistics for different augmentation strategies. \textit{T5} denotes the statistics of the original train set. \textit{+\ptrain, +\ptest, +\puniform} denote augmented datasets based on different PCFGs. We report the statsitics for both input sentence side and output meaning representation side. Thus, \textit{Avg length} under \textit{English} tab refers to the average length of input sentences. We report three statistics: average length (\textit{Avg length}), the coverage (expressed as a percentage) of bigrams in the test set by the training set (\textit{Bigrams}) and the coverage of entire instance in the test set by the training set (\textit{Instance}). 
% \todo{The percentage here is all 0 because I filtered  overlapped instances out.}
}
\label{tab:pcfg_stats }

\end{table*}
% The statistics of GeoQuery is present in Table \ref{tab:geoquery_pcfg_stats }. On the length split, \ptest covers more meaning representations than \ptrain and \puniform, and thus achieves the best performance. However, this pattern does not hold template split, where the improvement of \ptest is very limited and gets outperformed by \puniform. We do not have a good explanation for this. 

\paragraph{SCAN}

Table \ref{tab:scan_results} shows exact match accuracies on SCAN. We observe \ptest\ and \puniform\ substantially improve the performance on both splits, whereas the \ptrain\ only performs well on \textit{turnleft} split. All three strategies achieves the same performance on \textit{turnleft} split. This is because the meaning representation space of SCAN is too small and thus all possible meaning representations can be sampled by three strategies, which results in the same train set. The same case happens for \ptest\ and \puniform\ on the length split. Noticeably, our method outperforms GECA, which generates parallel data for data augmentation using templates. This suggests that sampling meaningful and useful meaning representations proves more effective than sampling limited parallel data in certain scenarios. 

% We also present the statistics of SCAN in Table \ref{tab:scan_pcfg_stats }. Both \ptest and \puniform covers most of output meaning representations in the test set, and thus achieve much higher performance than baseline and \ptrain.

\section{Discussion} \label{sec:discussion}

\begin{figure*}
  % \centering
  \begin{subfigure}{0.3\textwidth}
    \centering
    \includegraphics[width=\linewidth]{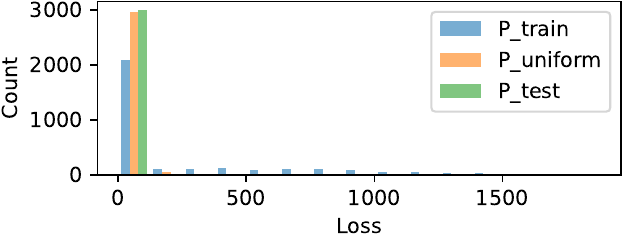}
    \caption{COGS structural types}
    \label{fig:loss:cogs}
  \end{subfigure}%
  \hfill
      \begin{subfigure}{0.3\textwidth}
    \centering
    \includegraphics[width=\linewidth]{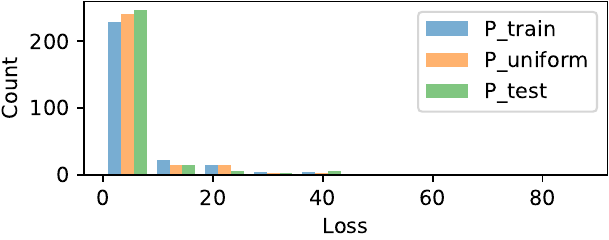}
    \caption{GeoQuery Template split}
    \label{fig:loss:geoquery}
  \end{subfigure}%
  \hfill
      \begin{subfigure}{0.3\textwidth}
    \centering
    \includegraphics[width=\linewidth]{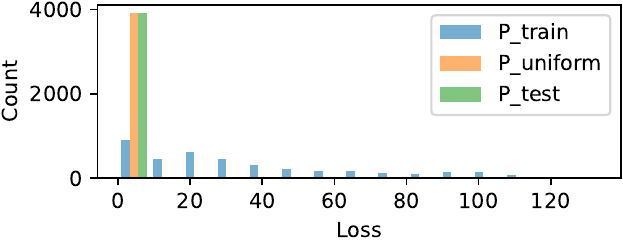}
    \caption{SCAN Length split}
    \label{fig:loss:scan}
  \end{subfigure}%
  \hfill
    \caption{Count of test instances with regard to different loss values.
    % \todo{replace \ptest with \puniform}
    }
    \label{fig:loss}
\end{figure*}

\begin{figure}
    \centering
  \begin{subfigure}{0.4\textwidth}
    \centering
    \includegraphics[width=\linewidth]{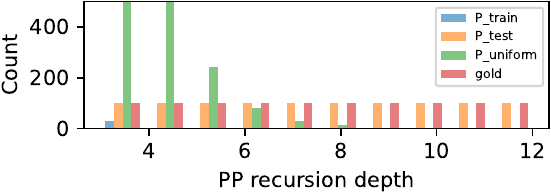}
    % \caption{COGS}
    \label{fig:cogs_pp_dist}
  \end{subfigure}%
  \hfill
    \begin{subfigure}{.4\textwidth}
    \centering
    \includegraphics[width=\linewidth]{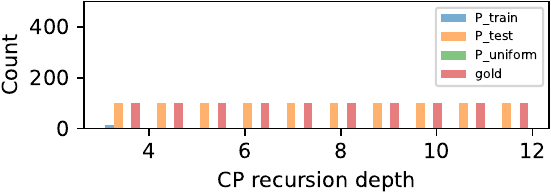}
    % \caption{COGS}
    \label{fig:cogs_cp_dist}
  \end{subfigure}%
  \caption{Depth distribution of train set for COGS.}
    \label{fig:depth_gen}
\end{figure}

The surprising finding so far is that across all four compositional generalization datasets, augmenting from \puniform\ performs on par with \ptest. This seems counterintuitive: the uniform augmentation strategy has no knowledge of the test data's distribution, and one would expect that augmentation data sampled from a grammar-based approximation to the test distribution should perform much better. We  therefore investigate this finding in detail.

\paragraph{Augmentation data statistics}

We present statistics of the generated augmentation data in Table \ref{tab:pcfg_stats }. For each corpus and augmentation method, we show the average sequence length, bigram coverage, and instance (i.e.\ exact sequence match) coverage for both input sentences and output MRs. 
The bigram coverage is determined by dividing the number of observed bigrams in the test set that also exist in the training set by the total count of possible bigrams in the test set.
Instance coverage is calculated analogously.

As expected, \ptest\ always yields the highest coverage values on the meaning representations, suggesting that the MR grammar approximates the test distribution effectively. On the other hand, instance-level coverage on the English side does not grow very high for any dataset. This indicates that the backtranslation model, which is trained on the original in-distribution data, still struggles to produce novel recombinations of the English sentences.

\puniform\ is on par with \ptest\ on many measures and datasets, and considerably outperforms \ptrain. This suggests that novel structural combinations are judged unlikely based on the training distribution, or are simply assigned a probability of zero because structures were entirely unobserved.

% We first observe that \ptest always give the highest test-set coverage of meaning representation instances and bigrams across all datasets, which suggests that our estimated grammar does approximate the test-set distribution. On the other hand, the coverage of input texts (i.e.\ \textit{Instance} numbers under \textit{Text}) does not grow much for all augmentation strategies. We consider this is because the backtranslation model trained on the in-distribution data still suffers from generalizing to novel combinations of natural language texts. Besides, \ptrain yields lower bigram coverage and shorter sequences compared to \ptest and \puniform on COGS, GeoQuery and SCAN. We consider this is because estimating parameters on train set leads to zero or very low probability for some specific grammar rules. This will make the sampling process ineffective and thus hard to cover the space of meaning representations.

It is remarkable that \ptest\ and \puniform\ produce meaning representations of similar length on COGS and could therefore be capable of generating augmentation data of similar structural complexity.
At the same time, \ptest\ achieves a significantly higher parsing accuracy on the PP and CP recursion generalization types. 
A plot of the distribution of the augmentation instances according to recursion depth (Fig.~\ref{fig:depth_gen}) reveals that while \ptest\ generates augmentation instances evenly across all recursion depths, \puniform\ emphasizes moderately (PP) or extremely (CP) shallow instances.\footnote{We hypothesize that the difference between PP and CP in the \puniform\ case is due to the fact that each level of CP recursion requires the use of two production rules, rather than just one for PP, making the generation of deeper structures comparatively less likely.}
This explains the difference in parsing accuracy, and further emphasizes that compositional generalization is not just challenging because transformers struggle when generalizing to longer inputs \cite{hupkes2020compositionality}, but also to structurally more complex inputs of similar length.

% . We find that \puniform yields instances with PP depth up to 8, which is consistent with its high accuracy for PP type in Table \ref{tab:cogs_results}. On the other hand, although the model observed long enough sequences from augmentation instances, only shallow CP recursions are observed, which leads to low accuracy on CP types. This suggests that the difficulty of compositional generalization not only comes from length generalization, but also depth generalization.  
% We can see \ptest covers almost all output ngrams in the test set and shares similar maximum structure depth, which leads to the best performance. \puniform shares the similar maximum output length with \ptest, but only contains instances with cp\_depth up to 2, which results in its low accuracy on CP recursions. \ptrain does not contain instances with larger depth, and thus does not improve the model .

\paragraph{Perplexity analysis}

We further investigate whether \puniform\ produces useful augmentation data simply because it produces arbitrary instances of higher complexity than \ptrain, or if \puniform\ actually models the test distribution in some way. To this end, we measure the perplexity of the meaning representations of the test set across four corpus variants under each model (Table~\ref{tab:perplexity}; see Appendix \ref{appendix:perplexity} for details).

We find that across three of the four datasets, \ptest\ and \puniform\ are close together, considerably outperforming \ptrain\ and the T5 baseline. An exception is CFQ, where the grammar introduces so much noise into the sampling process that all models are mostly on par. 
We consider this is because although \puniform\ has no particular knowledge of the test distribution built in, sampling from it covers enough MR n-grams that the test data becomes predictable. 

% Let's work this out in more detail later - AK
% \todo{I think this is a fair representation of your argument. I'm not sure I believe it -- isn't it really weird that uniform finds the test data so likely across all the corpora? Does this say something about the way the test sets were constructed?}
% \yk{I think this is because the rule space of meaning representation grammars are generally small and so uniformly sampling can cover it well.}

\begin{table}
\footnotesize
    \centering
    \begin{tabularx}{\linewidth}{lXXXX}
    \toprule
     & COGS & CFQ & GeoQuery & SCAN \\
     \cmidrule[0.5pt](l){2-5}
     % \cmidrule{}
     Models & & MCD1 & Template & Length \\
    \midrule
    T5 &  1.131 & 1.007 & 1.254 & 1.427 \\
    \quad+$P_{train}$ \dataaug & 1.133 & 1.005 & 1.252 & 1.124\\
    \quad+$P_{test}$ \dataaug & 1.001 & 1.005 & 1.166 & 1.006 \\
    \quad+$P_{uniform}$ \dataaug & 1.007 & 1.005 & 1.184 & 1.006\\
    \bottomrule
    \end{tabularx}
    \caption{Perplexity of models with different augmentation strategies on test set.  
    % \todo{plan to add results with additional MRs from dev set.} 
    % \todo{update numbers}
    }
    \label{tab:perplexity}
\end{table}

% \citet{csordas-etal-2021-devil} noticed that low perplexity alone does not directly entail high accuracy; 
% \todo{I don't understand this paragraph enough to rewrite it. The way you explain it, I think you are talking about the training loss going to zero for most instances; but in the discussion above we are talking about test loss/perplexity. Could you explain this a bit more clearly?}
% \yk{All "loss" I discussed in this paragraph refers to the loss on the test set. }
% \todo{Yeah okay, but I still don't understand it. Could you please rewrite this paragraph so it becomes clear (a) what the exact problem is and (b) why we care?}
% Despite the perplexity patterns above, it may not directly explain the higher accuracy of \ptest\ compared to \ptrain. According to \cite{csordas-etal-2021-devil}, 
% model accuracy on a compositional test set can increase simultaneously with the loss as training progresses. 
% This is because during training, the loss of most test instances decreases to zero but the loss of other instances spreads out and becomes exceptionally high. Is the higher perplexity of \ptrain\ compared to \ptest\ because of the same reason? To answer this question, we further plot the count of test-set instances with regard to the model loss on the test set in Figure \ref{fig:loss}. On all three datasets, \ptest\ yields a small loss on most test instances, while \ptrain\ yields large loss on some instances. This indicates that the lower perplexity of \ptest\ does translate into the higher accuracy on the test set. 

The increased perplexity of \ptrain\ in comparison to the other models is not evenly distributed across the test instances. In Fig.~\ref{fig:loss}, we plot a count of test instances for each loss value. Compared to \ptest\ and \puniform, the loss of \ptrain\ on some instances becomes exceptionally high, which results in higher perplexity and lower accuracy on such instances. Looking into the dataset, we find that such issue generally occurs on meaning representations with complex structures (e.g.\ deeper recursions for COGS and unseen program templates for GeoQuery). 
These structures are more predictable for models trained on \ptest\ and \puniform\ augmentation data, which contains such structures more frequently.

% Instead, such structures exist in meaning representations sampled from \ptest\ and \puniform\, which explains their lower perplexity on test set.  

\paragraph{Structure coverage}
% \paragraph{Breakdown performance improvements}
% \paragraph{The effect and limitation of augmentation} 

% \todo{I also haven't touched this paragraph. I think the general theme here is that we know that comp gen is hard because of unseen structures and unseen n-grams, and we want to check how well the augmentation strategies make these seen in the augmentation data. This is a point that I think we could make much more clearly with the pictures I suggested in the Teams chat. Let's see what we can come up with, and then revise the text here.}

According to \citet{bogin-etal-2022-unobserved}, a key feature that makes compositional generalization difficult is the presence of unobserved local structures (i.e.\ a connected sub-graph that occurs in the meaning representation) in the test set. Is the better performance and perplexity of \ptest\ and \puniform\ actually because they cover more structures in the test set? 

\begin{figure}
    \centering
    \begin{subfigure}{0.4\textwidth}
    \centering
    \includegraphics[width=\linewidth]{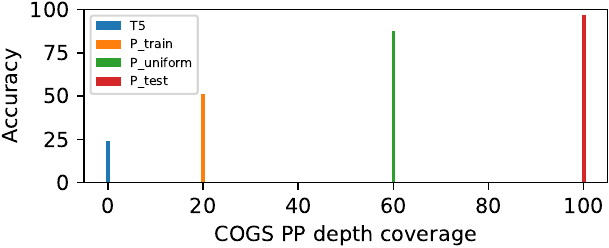}
    % \caption{Exact match accuracy of T5 on COGS generalization set with different maximum structure depths observed in the train set. }
    % \label{fig:acc_cogs_against_depth}
  \end{subfigure}%
  \hfill
    \begin{subfigure}{0.4\textwidth}
    \centering
    \includegraphics[width=\linewidth]{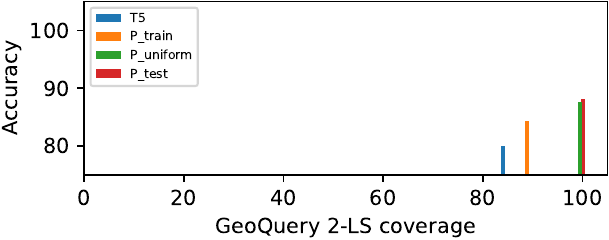}
    % \caption{Execution accuracy of T5 on GeoQuery test set (template split) with increasingly observed rule combinations in the train set.}
    % \label{fig:acc_geo_against_ls}
  \end{subfigure}
  \caption{Performance against the local structure coverage for different augmentation distributions.}
    \label{fig:acc_against_structure} 
\end{figure}

To answer this question, we further plot the accuracy of our models against the structure coverage on COGS and GeoQuery in Figure \ref{fig:acc_against_structure}. 
% Here ``structure coverage'' refers to the coverage of structures in the test set, under the same definition as in Table \ref{tab:pcfg_stats }. 
Here ``structure coverage'' refers to dividing the number of observed structure in the test set that also exist in the training set by the total count of possible structures in the test set.
For GeoQuery, we consider the template split and follow \citet{bogin-etal-2022-unobserved} in defining the local structure of a meaning representation as all pairs of parent nodes and their children in its parse tree (i.e.\ 2-LS). 
% For COGS, we only consider PP recursion type and use the maximum observed recursion depth to the represent the complexity of local structures. 
For COGS, we focus on the PP  recursion generalization type. Instead of considering local structures, we observe that the accuracy on such data is related to the maximal recursion depth observed in the train set. Thus we use PP recursion depth as a representative of global structures to calculate the structure coverage. 

Our results show that \ptest\ and \puniform\ yields a larger coverage of structures that occur in the test set than \ptrain. Furthermore, larger coverage is associated with higher accuracy. This  is consistent with \citet{bogin-etal-2022-unobserved}. Although \citet{gupta-etal-2022-structurally} and \citet{oren-etal-2021-finding} also show the benefit of introducing more complex structures into the train set, our results further suggest that synthesized meaning representations with back-translated sentences can still help. 
% We additionally report breakdown performance improvements by incrementally introducing these new local structures in Appendix \ref{appendix:breakdown}.

\paragraph{Qualitative error analysis} Finally, we conducted a qualitative analysis to identify specific cases in which our approach led to improvements. In Table \ref{tab:error_example}, the grammar rule \textit{River} $\rightarrow$ \textit{most River} is not observed by baseline and \ptrain, and thus the model struggles generating the bigram \textit{most river} corresponding to this rule, which leads to a large loss value (i.e.\ 26.1) for this instance. In contrast, \ptest\ covers all local structures, which allows the model to predict the instance correctly with a substantially lower loss (i.e.\ 0.1). 

\begin{table}
    \centering
    \footnotesize
    \begin{tabularx}{\linewidth}{lX}
    \toprule
    % \multicolumn{2}{c}{GeoQuery} \\
    % \midrule
     Input & what is the length of the river that runs through the most states ? \\
    \midrule 
    Gold & \lform{len \textbf{most river} traverse\_2 state all} \\
    \midrule
    T5  & \lform{len \red{intersection} riverid most state all} \\
    \midrule
    +\ptrain  & \lform{len \red{intersection} river traverse\_2 most state all} \\
    \midrule
    +\ptest  & \lform{len most river traverse\_2 state all} \\
    \bottomrule
    \end{tabularx}
    \caption{Examples from GeoQuery test set. 
    }
    \label{tab:error_example}
\end{table}

\paragraph{Sentence generation}

We report the performance of our backtranslation model in Table \ref{tab:bt_results}. Both exact match accuracy and BLEU score \cite{papineni-etal-2002-bleu} are used as evaluation metrics. All models achieve good BLEU scores, indicating the effectiveness of our backtranslation models. However, none of the model yields high accuracy,
% which is important for semantic parsing tasks since a meaning representation generally can only be mapped into the same sentence 
% \yk{This assumption actually does not hold on CFQ, GeoQuery and SCAN}
% . 
which suggests that our model can still learn to utilize such noisy data to achieve better performance. 

\begin{table}
\footnotesize
    \centering
    \begin{tabularx}{\linewidth}{lXXXX}
    \toprule
     & COGS & CFQ & GeoQuery & SCAN \\
     \cmidrule[0.5pt](l){2-5}
     % \cmidrule{}
     Metric & Struct & MCD1 & Template & Length \\
    \midrule
    Exact Match & 30.9 & 4.8 & 19.6 & 8.6 \\ 
    BLEU & 78.5 & 42.9 & 61.6 & 51.7 \\
    \bottomrule
    \end{tabularx}
    \caption{Results of our backtranslation model on the test sets for each task.  \textit{Struct} under \textit{COGS} means we only calculate the metric on structural generalization types.
    % \todo{plan to add results with additional MRs from dev set.} 
    % \todo{update numbers}
    }
    \label{tab:bt_results}
\end{table}

\section{Conclusion}

% In this paper we investigate the method proposed by \cite{wang-etal-2021-learning-synthesize} across different compositional generalization corpora. In contrast to their finding that estimating the grammar on train set works best, we show that a uniformly initialized grammar achieves better performance and serves as a better alternative of test distribution. Our conducted experiments show that this is because of the absence of structures in the generated data from grammar estimated towards training distribution. Our work show the potential of utilizing monolingual meaning representations for compositional generalization and also provides evidence of why such data helps. 

% In the future, it would be interesting to explore generating meaning representations in a more effective way. For example, one can generate such data from a large language model through prompting. Besides, it will be useful to investigate the impact of the quality of natural language sentences from backtranslation. 

We investigated the impact of the choice of augmentation distribution on compositional generalization. We found that a PCFG for the meaning representations with uniform rule weights supports much more effective data augmentation than one that is trained on the training data, and almost on part with one that is trained on the test data.
A detailed analysis revealed that this is because the uniform grammar both achieves low perplexity on the test meaning representations and greatly improves structural coverage.

Thus, sampling meaning representations from a uniform PCFG and backtranslating them into natural-language sentences can serve as a simple and efficient data augmentation strategy for compositional generalization. It would be interesting to investigate the space of augmentation distributions in more detail in future work to see, for instance, how the generation of structurally even more diverse augmentation instances can be encouraged.

% Our method can be described as sampling meaning representations from a probabilistic context-free grammar first and then backtranslate them into sentences. 
% We compared three distributions: training distribution, test distribution and uniform distribution and our results show that sampling from uniform distribution generally achieves better performance than training distribution.
% Our conducted experiments show that this is because uniform distribution yields better structure coverage than training distribution.
% Our work suggests that using uniformly sampled meaning representations for data augmentation can serve as a strong baseline for future compositional generalization works. 

\section{Limitations}
Our work assumes that the language of all possible meaning representations can be described with a context-free grammar, and that such a grammar is available or can be easily reconstructed by hand. Given that MRs are a formal language, this seems realistic, but can involve some manual effort. When the meaning representations are generated out of a knowledge base through a process that is not publicly accessible, such as in CFQ, hand-crafting a grammar for MRs can introduce noise.

In our evaluation, we use corpora that are either synthetic (COGS, CFQ, SCAN) or very small (GeoQuery). Thus, one should interpret conclusions on data augmentation for such corpora with care. To our knowledge, there are no compositional generalization datasets that use naturally occurring language. The robustness of our results across corpora still suggests the generality of our findings.

\bibliography{anthology,custom}
\bibliographystyle{acl_natbib}

\appendix

\section{Dataset details}
We report dataset statistics in Table \ref{tab:dataset_statistics}. COGS provides both an in-distributional test set (i.e.\ test) and an out-of-distributional test set (i.e.\ gen). For the splits of CFQ, GeoQuery and SCAN we used no in-distributional test set is provided.  
\begin{table}
    \centering
    \footnotesize
    \begin{tabular}{lcccc}
    \toprule
         Dataset & lr & batch\_size & weight\_decay & steps  \\
         \midrule
         COGS & 1e-5 & 2048 & 0 & 25k\\
         % \midrule
         CFQ & 7.4e-5 & 2048 & 1e-3 & 50k \\
         % \midrule
         GeoQuery & 1e-5 & 4096 & 1e-3 & 10k \\
         SCAN & 1e-5 & 1024 & 1e-3 & 25k \\
         \bottomrule
    \end{tabular}
    \caption{Hyperparmeters of baseline models used in our experiments. Batch size is quantified in terms of input tokens. 
    \textit{batch\_size} refers to the batch size during training. \textit{weight\_decay} refers to the weight decay used in the optimizer.
    \textit{lr} refers to the learning rate. \textit{steps} refers to the training steps we used to train the model. }
    \label{tab:hyperparmeters:baseline}
\end{table}

% \begin{table}
%     \centering
%     \footnotesize
%     \begin{tabular}{lcccc}
%     \toprule
%          & & Concat & \multicolumn{2}{c}{Pre-train} \\
%          Dataset & size & steps & steps1 & steps2  \\
%          \midrule
%          COGS & 21k & 25k & 25k &\\
%          % \midrule
%          CFQ & 100k & - & -  &\\
%          % \midrule
%          GeoQuery & 30k & 5k & 10k & \\
%          SCAN & 10k & 15k & 5k & \\
%          \bottomrule
%     \end{tabular}
%     \caption{Hyperparmeters of data augmentation methods used in our experiments. 
%     }
%     \label{tab:hyperparmeters:augmentation}
% \end{table}

\begin{table}
    \centering
    \footnotesize
    \begin{tabular}{lcc}
    \toprule
          & \multicolumn{2}{c}{Time (hours)}  \\
         Dataset & \textit{w.o.}\ aug & \textit{w.}\ aug \\
         \midrule
         COGS & 7 & 8 \\
         % \midrule
         CFQ & 10 & 10 \\
         % \midrule
         GeoQuery & 0.5 & 1\\
         SCAN & 20 & 4\\
         \bottomrule
    \end{tabular}
    \caption{Training time for our model on each dataset (1 run) in our experiments.}
    \label{tab:traintime}
\end{table}

\begin{table*}
    \centering
    \footnotesize
    \begin{tabular}{lccccccccc}
    \toprule
    Dataset & Split & \# train & \# dev. & \# test & \# gen & Vocab. size & Train len. & Test len. & Gen len. \\
    \midrule
    COGS & - & 24155 & 3000 & 3000 & 21000 & 809 & $22 / 48$ &$19/40 $ & $61 / 144$ \\
    \multirow{3}{*}{CFQ} & MCD1 & 95743 & 11968 & 11968 & - & 171 & $29/ 133$ & $30 / 103$ & - \\
    & MCD2 & 95743 & 11968 & 11968 & - & 171 & $29/ 133$ & $30 / 103$ & - \\
    & MCD3 & 95743 & 11968 & 11968 & - & 171 & $29/ 133$ & $30 / 103$ & - \\
    \multirow{2}{*}{GeoQuery}& template & 544 & 60 & 276 & - & 324 & $23/17$ & $19/12$ & - \\
     & length  & 540 & 60 & 280 & - & 324 & $13/7$ & $23/17$ & - \\
    \multirow{2}{*}{SCAN} & Turnleft & 21890 & - & 1208 & - & 19 & $9/48$ & $8/27$ & - \\
     & length  & 16990 & - & 3920 & - & 19 & $9/22$ & $9/48$ & - \\
     
    \bottomrule
    \end{tabular}
    \caption{Statistics for all our datasets. \# denotes the number of instances in the dataset. Vocab.size denotes the size of vocabulary for the dataset, which consists of input tokens and output tokens. Train.len denotes the maximum length of the input tokens and output tokens in the train set. Test.len and Gen.len denote the maximum length in the test and generalization set.}
    \label{tab:dataset_statistics}
\end{table*}

% \begin{table*}[!htbp]
%     \centering
%     \footnotesize
%     \begin{tabular}{lccccccccc}
%     \toprule
%     Dataset & Split & \# train & \# dev. & \# test & \# gen & Vocab. size & Train len. & Test len. & Gen len. \\
%     \midrule
%     COGS & - & 42675 & 3000 & 3000 & 21000 & 1424 & $46 / 140$ &$19/40 $ & $61 / 144$ \\
%     \multirow{3}{*}{CFQ} & MCD1 & 107711 & 11968 & 11968 & - & 176 & $35/ 133$ & $30 / 103$ & - \\
%     & MCD2 & 107711 & 11968 & 11968 & - & 176 & $32/ 133$ & $30 / 103$ & - \\
%     & MCD3 & 107711 & 11968 & 11968 & - & 174 & $33/ 133$ & $30 / 103$ & - \\
%     \multirow{2}{*}{GeoQuery \footnotesize (FunQL)} & template & 30000 & 60 & 276 & - & 568 & $28/137$ & $19/12$ & - \\
%      & length  & 30000 & 60 & 280 & - & 421 & $18/100$ & $23/17$ & - \\
%      \multirow{2}{*}{SCAN} & Turnleft & 30000 & 60 & 276 & - & 568 & $28/137$ & $19/12$ & - \\
%      & length  & 30000 & 60 & 280 & - & 421 & $18/100$ & $23/17$ & - \\
%     \bottomrule
%     \end{tabular}
%     \caption{Statistics for augmented datasets used in our experiments.}
%     \label{tab:aug_dataset_statistics}
% \end{table*}

\section{Training details} \label{appendix:training}

\subsection{Evaluation metrics}
For all tasks, we report exact match accuracy of our model, which means that the output sequence is correct only if each output token is correctly predicted. For GeoQuery, we additionally report execution accuracy, which means we execute generated FunQL code and calculate the accuracy of the outputs. This metric can better measure the generalization ability of our model since one input sentence can be mapped into multiple correct FunQl queries. For example, \textit{how long is the rio grande river} can be parsed into either \textit{answer ( len ( river ( riverid ( rio grande ) ) ) )} or \textit{answer ( len ( intersection ( riverid ( rio grande ),  ( river ( all ) ) ) ) )}. Both queries return the correct value. We use the code from \cite{herzig-berant-2021-span-based} to calculate the execution accuracy.

\subsection{Hyperparameters}
\paragraph{Baseline.} We use \textit{t5-base}\footnote{\url{https://huggingface.co/t5-base}} (220 million parameters) as our baseline for all experiments. We use the default subword vocabulary and do not extend it with new words. We use Adam \cite{DBLP:journals/corr/KingmaB14} as our optimizer. Since \cite{csordas-etal-2021-devil} shows that early stopping based on in-distribution validation set leads to low performance on out-of-distribution test set, we do not apply early stopping for COGS, GeoQuery and SCAN and only use the checkpoint at the end of training, following \cite{herzig-etal-2021-unlocking}. CFQ provides out-of-distribution development set, so we use exact match accuracy on the development set as the validation metric.
No learning rate scheduler is used for all experiments.
During evaluation, we use beam search with beam size 4.
% \cite{csordas-etal-2021-devil} shows that early stopping based on in-distribution validation set leads to . 
Task-specific hyperparameters are present in Table \ref{tab:hyperparmeters:baseline}. 

We only perform hyperperameter selection for the learning rate hyperparameter. For CFQ, we perform random search to select the best learning rate based on the model accuracy on the dev set. The search space for the learning rate is (1e-5, 1e-3). For other datasets where the development set is not distributed the same as the test set (i.e.\ COGS and GeoQuery) or the development set is not provided (i.e.\ SCAN), we sample a held-out development set from its test set following \cite{zheng-lapata-2022-disentangled} for hyperparameter selection. The best learning rate is selected from values in \{1e-4, 1e-5\} based on the model accuracy on the sampled development set.

\paragraph{Data augmentation.} We maximally sample 21k, 100k, 30k and 10k unique meaning representations for COGS, CFQ, GeoQuery and SCAN respectively. For SCAN, we find that the size of possible meaning representations is small (i.e.\ 9228 unique meaning representations) and thus we sample all possible unique meaning representations from our PCFG.

% \subsection{preprocessing}
\subsection{Other details}
We use Allennlp \cite{gardner-etal-2018-allennlp} for our implementation. Experiments are run on Tesla
A100 GPU cards (80GB). Table \ref{tab:traintime} shows the training time cost. 
% The training time on the augmented train set of SCAN is smaller than that on the original train set. This is because the augmented data helps the model converge on the development set faster and .  

\section{Detailed results} \label{appendix:detail_results}
We report detailed experimental results in Table \ref{tab:detailed_results}. Both means and standard deviations are reported over 5 runs for each model. As discussed in Section \ref{sec:method}, we experiment with two ways to utilize synthesized data: concatenating them with the original train set and pretrain the model on them first and then finetune on the original train set. We report numbers for both settings, with \textit{Concat} refers to concatenation and \textit{Pretrain} refers to pretraining the model first and then fine-tuning it. 

\section{Grammar details} \label{appendix:grammar}
We use PCFG provided in \citet{kim-linzen-2020-cogs} for COGS and hand-written grammars for CFQ, GeoQuery and SCAN. In this section, we mainly introduce details of our used grammar for these three hand-written grammars. 
% We focus on COGS and GeoQuery. Although \cite{keysers-etal-2020-measuring} also has a grammar to generate CFQ meaning representations, the  
\subsection{Grammar design}
\paragraph{GeoQuery} 
The meaning represnetation of GeoQuery is based on FunQL. Following the definations of FunQL \footnote{\url{https://www.cs.utexas.edu/~ml/wasp/geo-funql.html}}, we can easily write a context-free grammar for it.  We adopted the FunQL grammar used in \cite{guo-etal-2020-benchmarking-meaning} and extends it with some rules to fit our dataset.
A selection of our context-free grammar rules are shown in Table \ref{tab:grammar:geo}. 

\begin{table}
    \centering
    \footnotesize
    \begin{tabular}{l}
    \toprule
         S $\rightarrow$ answer ( Var ) \\
         Var $\rightarrow$ City \\
        Var $\rightarrow$ Place \\

        Var $\rightarrow$ State\\

        City $\rightarrow$ CityNonterm\\

        City $\rightarrow$ CityTerm\\

        CityNonterm $\rightarrow$ city ( City )\\

        CityNonterm $\rightarrow$ loc\_2 ( State )\\
        
        CityTerm $\rightarrow$ city ( all ) \\

        CityTerm $\rightarrow$ capital ( all ) \\

    \bottomrule
    \end{tabular}
    \caption{Part of our FunQL grammar.}
    \label{tab:grammar:geo}
\end{table}

\paragraph{SCAN} SCAN is a synthetic dataset generated by \citet{lake-baroni-2018-generalization}. They generate the dataset by generating commands (i.e.\ input sentences) first and then translating commands into action sequences (i.e.\ meaning representations) with a translation function. Instead, we write a context-free grammar for meaning representations. A selection of our context-free grammar rules are shown in Table \ref{tab:grammar:scan}.

\begin{table}
    \centering
    \footnotesize
    \begin{tabular}{l}
    \toprule
S $\rightarrow$ Command \\

% S $\rightarrow$ Command Command\\

Command $\rightarrow$ Walk\_command\\

% Command $\rightarrow$ look\_command\\

% Command $\rightarrow$ jump\_command\\

Walk\_command $\rightarrow$ Walk\_actions \\

Walk\_actions $\rightarrow$ LWalk \\
LWalk $\rightarrow$ Turn\_left Walk \\
Turn\_left $\rightarrow$ i\_turn\_left \\
Walk $\rightarrow$ i\_walk \\
    \bottomrule
    \end{tabular}
    \caption{Part of our SCAN grammar.}
    \label{tab:grammar:scan}
\end{table}

\begin{table}
    \centering
    \footnotesize
    \begin{tabular}{l}
    \toprule
S $\rightarrow$ Prefix Main \\

Main $\rightarrow$ lb Conjuncts rb\\

Conjuncts $\rightarrow$ Conjuncts . Conjunct\\

Conjuncts $\rightarrow$ Conjunct\\

Conjunct $\rightarrow$ Unary\_relation\\

Unary\_relation $\rightarrow$ ( Var a Film\_unary\_arg )\\

Film\_unary\_arg $\rightarrow$ film.film \\

Var  $\rightarrow$ M0 \\

    \bottomrule
    \end{tabular}
    \caption{Part of our SPARQL grammar.}
    \label{tab:grammar:cfq}
\end{table}

\begin{figure*}
  \centering
  \begin{subfigure}{0.5\textwidth}
    \centering
    \includegraphics[width=\linewidth]{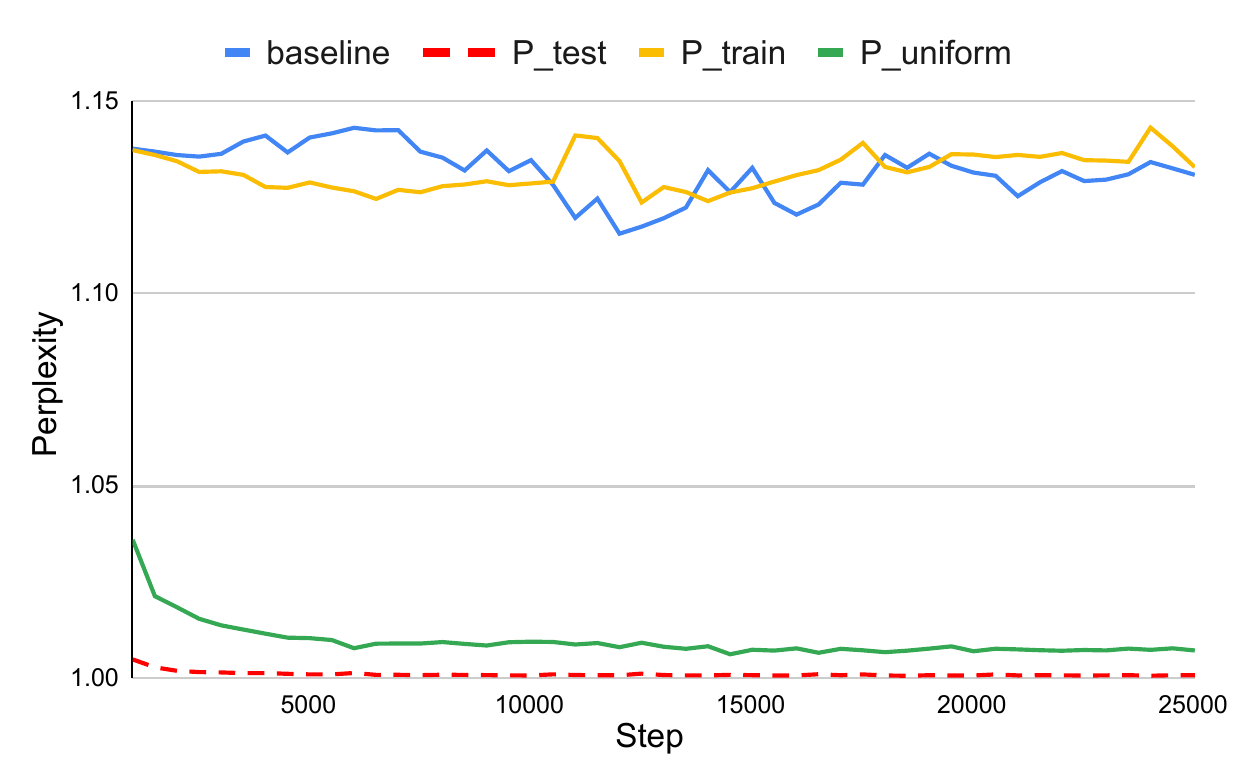}
    \caption{COGS}
    \label{fig:ppl:cogs}
  \end{subfigure}%
  \hfill
    \begin{subfigure}{0.5\textwidth}
    \centering
    \includegraphics[width=\linewidth]{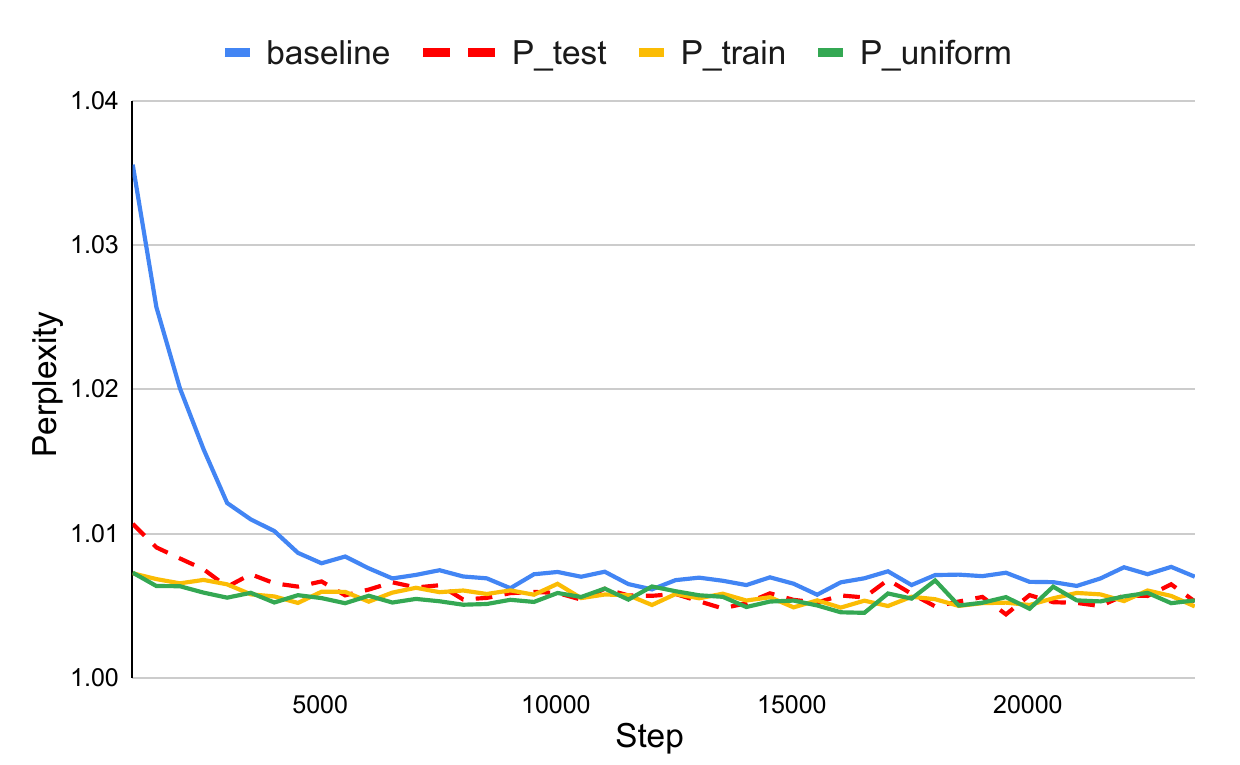}
    \caption{CFQ MCD1}
    \label{fig:ppl:cfq}
  \end{subfigure}%
  \hfill
      \begin{subfigure}{0.5\textwidth}
    \centering
    \includegraphics[width=\linewidth]{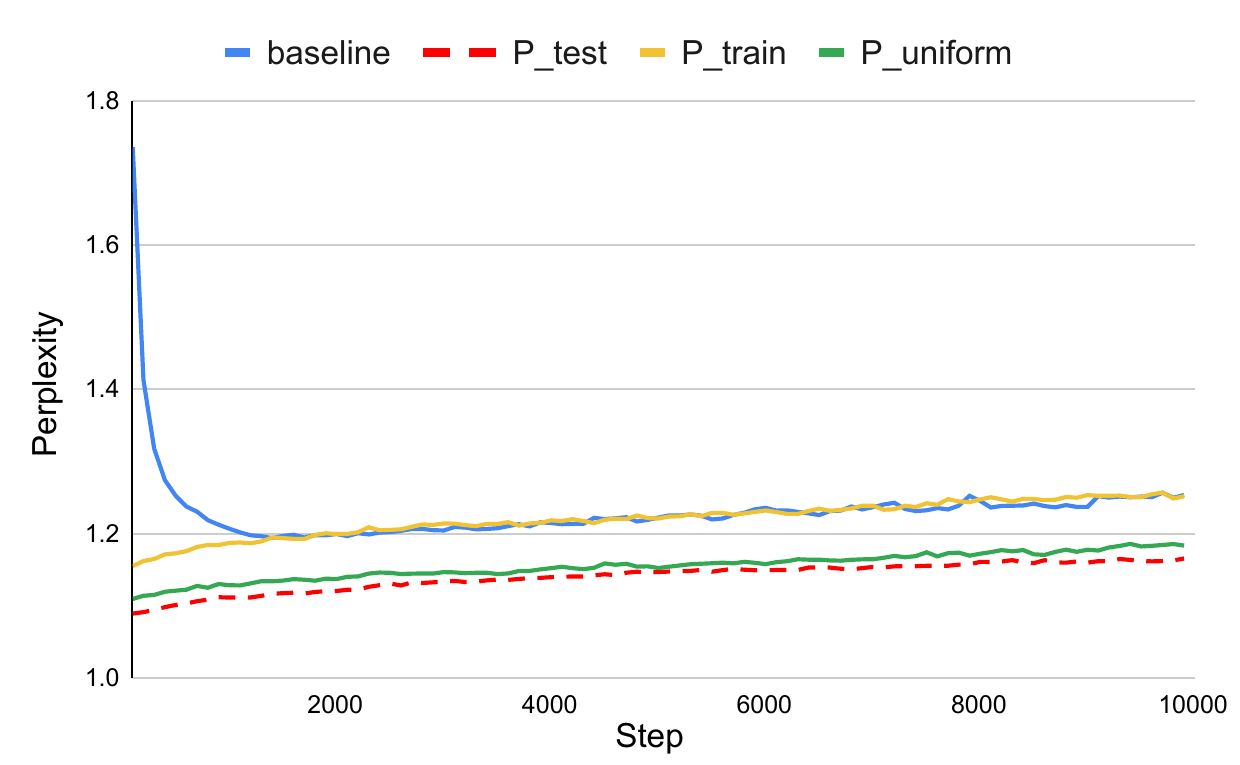}
    \caption{GeoQuery Template split}
    \label{fig:ppl:geoquery}
  \end{subfigure}%
  \hfill
      \begin{subfigure}{0.5\textwidth}
    \centering
    \includegraphics[width=\linewidth]{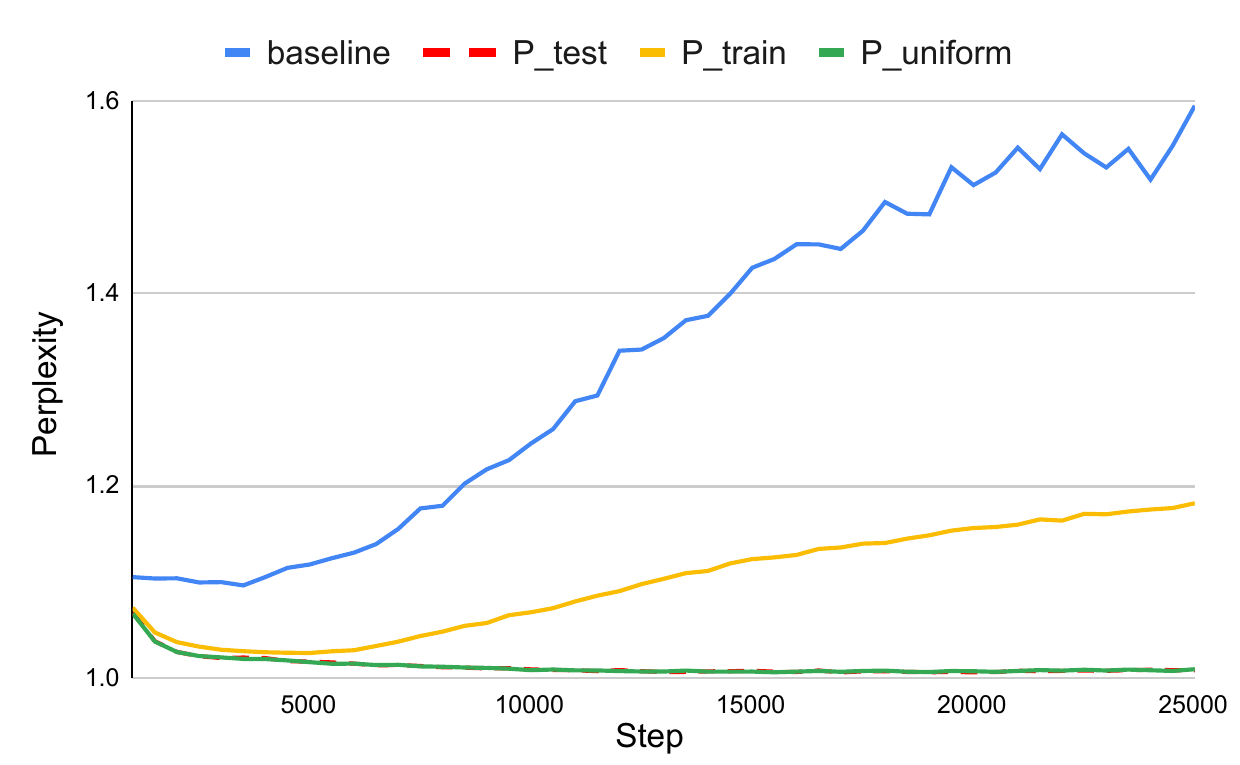}
    \caption{SCAN Length split}
    \label{fig:ppl:scan}
  \end{subfigure}%
  \hfill
  \caption{Perplexity of models with different augmentation strategies on test set. The x-axis refers to training steps.
  % \todo{On CFQ and GeoQuery, the curve for augmented models start from a low perplexity. This is because for these two datasets we pretrain the model to utilize augmented data. For COGS and SCAN, we directly concatenate the augmented data with the train set. I will run experiments with both pretraining and concatenation on all datasets to complete the picture. }
  }
  \label{fig:ppl}
\end{figure*}

\begin{figure}
    \centering
    \begin{subfigure}{0.4\textwidth}
    \centering
    \includegraphics[width=\linewidth]{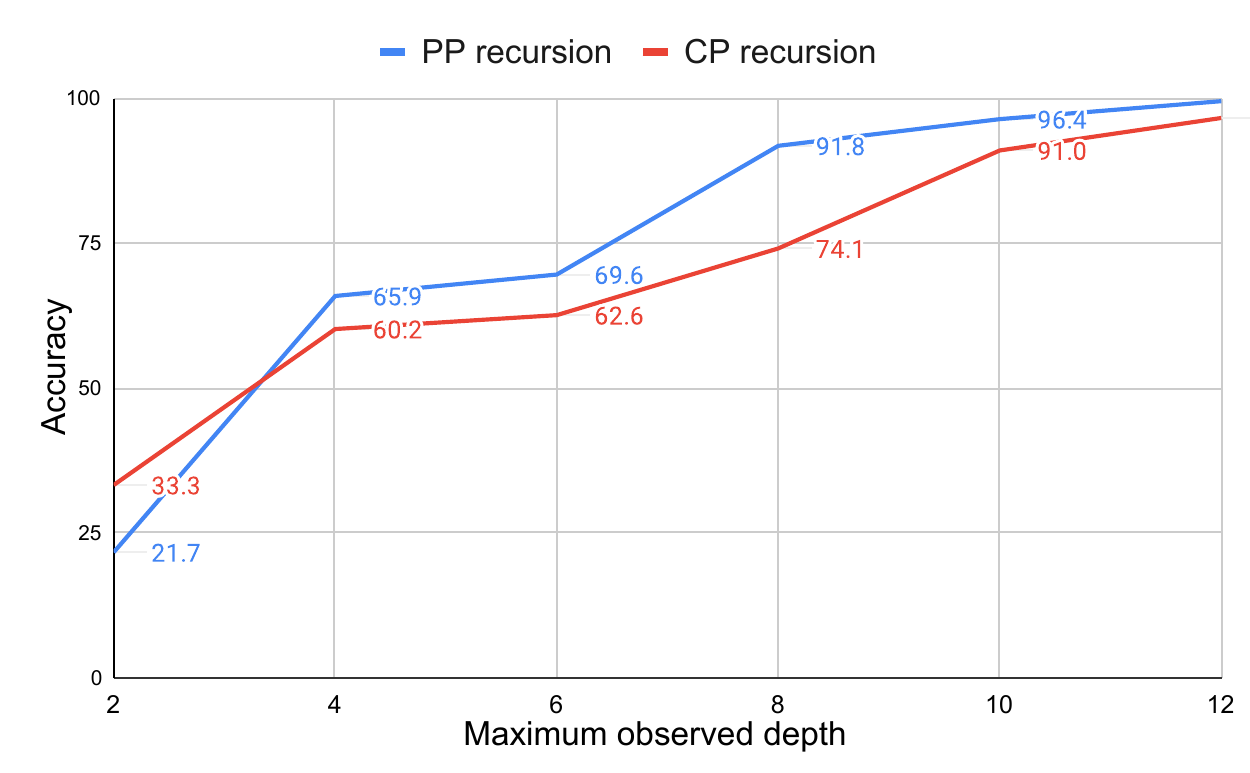}
    \caption{Exact match accuracy of T5 on COGS generalization set with different maximum structure depths observed in the train set. }
    \label{fig:acc_cogs_by_depth}
  \end{subfigure}%
  \hfill
    \begin{subfigure}{0.4\textwidth}
    \centering
    \includegraphics[width=\linewidth]{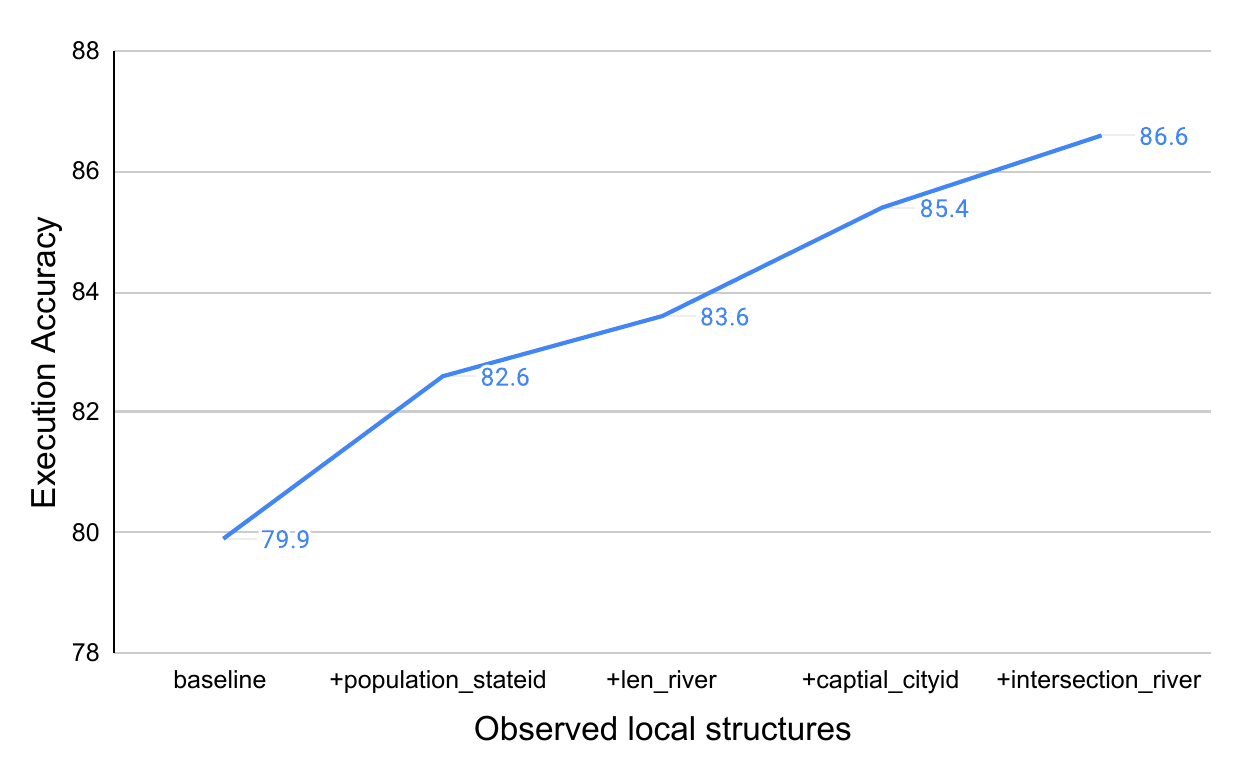}
    \caption{Execution accuracy of T5 on GeoQuery test set (template split) with increasingly observed rule combinations in the train set.}
    \label{fig:acc_geo_by_ls}
  \end{subfigure}
  \caption{Performance curve with regard to train sets with incrementally added structures.}
    \label{fig:acc_by_structure} 
\end{figure}

\paragraph{CFQ} CFQ is a synthetic dataset generated by \citet{keysers-etal-2020-measuring}. They generate the natural language sentences and corresponding intermediate logical forms first, and then apply multiple rules to obtain the SPARQL meaning representations. Designing a context-free grammar for SPARQL is hard because it contains variables and each relation only accepts specific typed variables as arguments. For example, the object of \textit{film.writer.film} relation should be a film. 
% To address such type constraints, we first replace all variable strings in a meaning representation with a special token \textit{var} and then only construct the grammar for the anonymized meaning representation. 
In our grammar, we consider all variable strings are produced by  the nonterminal \textit{Var} and we do post-process to filter out samples that do not follow type constraints described above. 
A selection of our context-free grammar rules are shown in Table \ref{tab:grammar:cfq}.

In our experiments, we find this setting still generates most noisy meaning representations due to redundant conjuncts (e.g.\ \textit{SELECT DISTINCT ?x0 WHERE \{ ( FILTER ( ?x0 != M0 ) ) .\ ( M5 ( film.editor.film ) ( ?x1 ) ) \}}). A better solution might be to construct the PCFG based on the graph structure.  

\subsection{Parameter estimation}
To estimate parameters of a grammar on a dataset based on maximum likelihood estimation, we first parse meaning representations in the dataset with our grammar rules described above. We implement this with NLTK package\footnote{\url{https://www.nltk.org/}}. We binarize our grammar rules to adopt parsing methods in NLTK. 

\paragraph{Ambiguous trees} For CFQ, all meaning representations can be parsed into unambiguous trees. For GeoQuery and SCAN, parsing results in ambiguous trees for some cases. For a meaning representation with $N$ ambiguous parse trees, we simply use a count $1/N$ as the count for rules in each tree to estimate their parameters.

% \paragraph{Variables in SPARQL} For CFQ, we only consider the anonymized meaning representations and independently estimate the probability of \textit{var} being mapped to a specific variable \textit{M0}). 
% We then use NLTK package\footnote{\url{https://www.nltk.org/}} to create our PCFG and sample from it.

\section{Additional experiments}

\subsection{Perplexity curve} \label{appendix:perplexity}

We plot the perplexity curve of different models on test set for each task in Figure \ref{fig:ppl}. For CFQ and CFQ, the perplexity at the beginning is already very small. This is because on these two datasets we pretrain the model on synthesized data first, since direct concatenating the synthesized data only hurts the performance. We can observe that for COGS, GeoQuery and SCAN, the perplexity of \ptest\ always achieves the lowest perplexity and \puniform\ gives lower perplexity than \ptrain. On CFQ, all three augmentation distributions achieves lower perplexity than baseline T5 and performs on par. This pattern holds during the entire training process, which serves as further evidence for the discussion in Section \ref{sec:discussion}.

\subsection{Breakdown performance improvements} \label{appendix:breakdown}

We also conduct a more detailed analysis to investigate how the performances evolve as more complex structures get observed. Specifically, we address the PP and CP recursion generalization types on COGS and GeoQuery template split. For COGS, we incrementally augment the train set with more complex data (i.e.\ deeper recursions) in increments of 100 instances per depth. For GeoQuery, we manually select four local structures \textit{population\_1 stateid}, \textit{len river}, \textit{capital cityid}, \textit{intersection river} that pose challenges for the baseline parser's predictions yet are present in the \ptest\ set. We incrementally introduce each pattern into the train set. 
As shown in Figure \ref{fig:acc_by_structure}, as more complex MR structures being observed by the model, its performance gets better improved. 

% Our analysis reveals that the model performance grows as the complexity of augmented data increases. This indicates that incorporating complex structures indeed improves the accuracy.

\subsection{Error examples in augmented data}

Another limitation of our data augmentation method is that our backtranslation model do not achieve high accuracy on out-of-distribution test data (Table \ref{tab:bt_results}), which thus introduces noise into the augmented training data. 
We present error examples in the augmented training data in Table \ref{tab:bt_errors}. 
On COGS, the backtranslation model tends to generate sentences with seen linguistic structures in the training data (e.g.\ PP dative: \textit{called the table to}) instead of unseen structures (e.g.\ Subject PP: \textit{A teacher on the table}). 
Also, given a meaning representation with deep recursion structures, the model may ignore some structures (e.g.\ \textit{in the bottle} ) and not translate them. 
% Similar patterns can also be observed in GeoQuery (e.g.\ \textit{river traverses} is not translated) and SCAN (e.g.\ \textit{run around right thrice} is never observed in the training set of SCAN length split because it will result in a long MR.)
Similar patterns can also be observed in GeoQuery and SCAN:
\textit{what state has the most cities} is an observed sentence in the training data of the GeoQuery backtranslation model, and thus the model tends to translate the given MR into this sentence ignoring the structure \textit{river traverse\_2};
\textit{run around right thrice} is never been observed in the training set of the SCAN backtranslation model, and thus the backtranslated sentence cannot generate such structures. 
On CFQ, we further observe that the model may generate additional phrases whose meaning is not present in the MR (e.g. \textit{was written by M5}). 
\begin{table*}[htb!]

    \footnotesize
    \setlength{\tabcolsep}{1.5pt}
    \begin{tabularx}{\linewidth}{lX}
    \toprule
    \multicolumn{2}{c}{COGS} \\
    \midrule
    MR &  call ( agent = teacher ( nmod . on = * table ) , theme = Emma ) \\
    \midrule
    Backtranslation & A teacher \red{called the table to} Emma .  \\
    \midrule
    Annotated & A teacher on the table called Emma . \\
    \midrule
    MR &  offer ( agent = * cat , theme = * block ( nmod . in = house ( nmod . on = towel ( nmod . in = tin ( nmod . in = * bottle ( nmod . in = car ( nmod . beside = * corpse ( nmod . on = * canvas ( nmod . beside = * bed ( nmod . beside = table ( nmod . in = * bag ( nmod . in = * hole ) ) ) ) ) ) ) ) ) ) ) )  \\
    \midrule
    Backtranslation & The cat offered the block in a house on a towel in a tin \red{in a car} beside the corpse on the canvas beside the bed beside a table in the bag in the hole .   \\
    \midrule
    Annotated & The cat offered the block in a house on a towel in a tin in the bottle in a car beside the corpse on the canvas beside the bed beside a table in the bag in the hole.  \\
    \midrule
        \multicolumn{2}{c}{CFQ} \\
    \midrule
    MR & SELECT DISTINCT ?x0 WHERE \{ ( ?x0 ( film.film.prequel , film.film.sequel ) ( M5 ) ) \} \\
    \midrule
    Backtranslation & What prequel and sequel of M5 \red{was written by M5} \\
    \midrule
    Annotated & What is the prequel and sequel of M5 \\
    \midrule
        \multicolumn{2}{c}{GeoQuery (template split)} \\
    \midrule
    MR &  answer most river traverse\_2 most state loc\_1 city all \\
    \midrule
    Backtranslation & what \red{state has the most cities} \\
    \midrule
    Annotated & what river traverses the state that has the most cities \\
    \midrule
        \multicolumn{2}{c}{SCAN (length split)} \\
    \midrule
    MR & RTURN RUN RTURN RUN RTURN RUN RTURN RUN RTURN RUN RTURN RUN RTURN RUN RTURN RUN RTURN RUN RTURN RUN RTURN RUN RTURN RUN RUN RUN RUN \\
    \midrule
    Backtranslation & run around right \red{twice} and run thrice \\
    \midrule
    Annotated & run around right thrice and run thrice \\
    \bottomrule
    \end{tabularx}
    \caption{Examples of incorrect training data introduced by backtranslation. \textit{MR} refers to the meaning representation sampled from the grammar. \textit{Backtranslation} refers to the corresponding English sentence generated by the backtranslation model. \textit{Annotated} refers to the human labeled English sentence for the MR. }
    \label{tab:bt_errors}
\end{table*}

In our experiments, we did not explore other hyperparameters (e.g.\ learning rate) for the backtranslation model and directly adapted the same setting from our semantic parser. However, one would expect that improving the performance of the backtranslation model can increase the benefit of our method if we assume our sampled meaning representations can cover most patterns in the test set. 
Future research could look into better sampling methods for backtranslated English sentences \cite{edunov-etal-2018-understanding}, or converting the input meaning representation into more text-similar representations \cite{mehta-etal-2022-improving}. Compositional generalization in MR-to-text generation tasks, which is still an underexplored area, is also an interesting direction for future research.

\begin{table*}[htbp]
    \centering
    \footnotesize
    % \resizebox{\linewidt÷h}{!}{
    \begin{tabular}{ll|ccc|c}
    \toprule
     & & \multicolumn{4}{c}{COGS} \\
    \multicolumn{2}{c}{Model} & Obj to Subj PP & CP recursion & PP recursion & Overall \\
    \midrule
    & T5 & $88.2 \pm 3.6$ & $32.3 \pm 3.7$ & $24.1 \pm 6.4$ & $91.0 \pm 0.5$ \\
    \multirow{3}{*}{Concat} & +\ptrain & $89.4 \pm 2.3$ & $43.5 \pm 8.7$ & $51.2 \pm 7.5$  & $92.9 \pm 0.9$ \\
    & +\ptest  & $94.6 \pm 0.1$ & $95.7 \pm 2.8$ & $96.7 \pm 5.0$ & $99.3 \pm 0.4$ \\
    & +\puniform & $94.8 \pm 0.0$& $50.7 \pm 2.4 $& $87.7 \pm 1.0 $ & $95.9 \pm 0.1$\\
    \multirow{3}{*}{Pretrain} & +\ptrain & $85.8 \pm 6.5$ & $41.3 \pm 11.3$ & $51.6 \pm 5.8$ & $92.8 \pm 0.6$ \\
    & +\ptest  & $94.8 \pm 0.0$& $43.1 \pm 5.7 $& $85.0 \pm 4.0 $ & $99.2 \pm 0.2$ \\
    & +\puniform & $94.6 \pm 0.1$ & $94.8 \pm 0.2$ & $92.7 \pm 4.9$ & $95.6 \pm 0.5$\\
    \midrule
    & & \multicolumn{4}{c}{CFQ} \\
    \multicolumn{2}{c}{Model} & MCD1 & MCD2 & MCD3 & Average \\
    \midrule
    & T5 & $89.8 \pm 0.8$ & $74.7 \pm 1.8$ & $74.0 \pm 0.9$ & $79.4 \pm 2.4$ \\
    \multirow{3}{*}{Concat} & +\ptrain & $49.5 \pm 1.9$ & $47.1 \pm 1.3$ & $51.2 \pm 2.9$  & $49.2 \pm 1.0$ \\
    & +\ptest  & $39.0 \pm 1.3$ & $44.7 \pm 2.3$ & $42.3 \pm 0.7$ & $42.0 \pm 0.9$ \\
    & +\puniform & $57.5 \pm 4.1$& $59.4 \pm 2.4 $& $55.2 \pm 3.3 $ & $57.4 \pm 1.8$\\
    \multirow{3}{*}{Pretrain} & +\ptrain & $89.9 \pm 1.2$ & $77.9 \pm 2.9$ & $75.8 \pm 1.0$  & $81.2 \pm 2.1$ \\
    & +\ptest  & $90.4 \pm 0.7$ & $79.1 \pm 1.7$ & $75.5 \pm 2.7$ & $81.7 \pm 3.6$ \\
    & +\puniform & $91.2 \pm 1.1$& $78.8 \pm 1.7 $& $74.3 \pm 1.7 $ & $81.4 \pm 1.1$\\
        \midrule
    & & \multicolumn{4}{c}{GeoQuery} \\
    \multicolumn{2}{c}{Model} & Template & Length &  &  \\
    \midrule
    & T5 & $73.9 \pm 2.6$ & $46.1 \pm 1.5$ & & \\
    \multirow{3}{*}{Concat} & +\ptrain & $39.0 \pm 0.9$ & $20.7 \pm 0.6 $& & \\
    & +\ptest  & $52.3 \pm 1.3$ & $35.6 \pm 1.7$ & & \\
    & +\puniform & $22.4 \pm 1.7$& $5.1 \pm 0.3 $&  &\\
    \multirow{3}{*}{Pretrain} & +\ptrain & $74.1 \pm 1.6$ & $56.1 \pm 2.1 $& & \\
    & +\ptest  & $80.1 \pm 1.7$ & $60.4 \pm 2.4$ & & \\
    & +\puniform & $79.3 \pm 1.3$& $60.1 \pm 0.6 $&  &\\
        \midrule
    & & \multicolumn{4}{c}{SCAN} \\
    \multicolumn{2}{c}{Model} & Turnleft & Length &  &  \\
    \midrule
    & T5 & $73.9 \pm 2.6$ & $4.4 \pm 0.9$ & & \\
    \multirow{3}{*}{Concat} & +\ptrain & $92.9 \pm 14.4$ & $8.1 \pm 1.3 $& & \\
    & +\ptest  & $92.9 \pm 14.4$ & $60.5 \pm 2.5 $ & & \\
    & +\puniform & $92.9 \pm 14.4$ & $60.5 \pm 2.5 $ &  &\\
    \multirow{3}{*}{Pretrain} & +\ptrain & $75.5 \pm 5.4$ & $15.5 \pm 1.5 $& & \\
    & +\ptest  & $75.5 \pm 5.4$ & $15.9 \pm 1.3$ & & \\
    & +\puniform & $75.5 \pm 5.4$ & $15.9 \pm 1.3 $&  &\\
    \bottomrule
    \end{tabular}
    % }
    \caption{Detailed results in our experiments.}
    \label{tab:detailed_results}
\end{table*}

\end{document}